\documentclass[11pt]{article}

\usepackage[letterpaper,lmargin=1.5in,rmargin=1.5in,tmargin=1in,bmargin=1in]{geometry}
\usepackage[utf8]{inputenc}
\usepackage[T1]{fontenc}
\usepackage{libertine}
\usepackage{amsmath}
\usepackage{amsthm}
\usepackage{amsfonts}
\usepackage{amssymb}
\usepackage{microtype}
\usepackage{graphicx}
\usepackage{subcaption}
\usepackage{booktabs}
\usepackage{nicefrac}
\usepackage{xcolor}
\usepackage{float}
\usepackage[round,authoryear]{natbib}
\usepackage[shortlabels]{enumitem}
\setlist{topsep=1pt,parsep=1pt,partopsep=1pt,itemsep=1pt}
\usepackage{hyperref}
\definecolor{mydarkblue}{rgb}{0,0.08,0.45}
\hypersetup{
  colorlinks=true,
  linkcolor=mydarkblue,
  citecolor=mydarkblue,
  filecolor=mydarkblue,
  urlcolor=mydarkblue,
  pdfborder=0 0 0,
  pdfview=FitH
}

\title{Transformers Learn Two Competing Sorting Algorithms,\\and Only One Length-Generalizes}

\newif\ifshowauthornotes
\showauthornotestrue % Set to \showauthornotesfalse to hide collaborator notes.
\newcommand{\aff}[1]{{\small\itshape #1}}
\newcommand{\nathan}[1]{\ifshowauthornotes {\color{blue}\textbf{NATHAN:} #1}\fi}

\author{%
  \begin{tabular}[t]{c}
    Khashayar Gatmiry\thanks{Equal contribution. Author ordering determined by a coin flip.} \\
    \aff{UC Berkeley \& Apple}
  \end{tabular}
  \hspace{1em}
  \begin{tabular}[t]{c}
    Nathan W.\ Henry\footnotemark[1] \\
    \aff{UC Berkeley}
  \end{tabular}
  \hspace{1em}
  \begin{tabular}[t]{c}
    Chloe H.\ Su \\
    \aff{Kempner Institute,} \\
    \aff{Harvard University}
  \end{tabular}
  \\[3em]
  \begin{tabular}[t]{c}
    Emmanuel Abbe \\
    \aff{EPFL, Apple}
  \end{tabular}
  \hspace{1em}
  \begin{tabular}[t]{c}
    Samy Bengio \\
    \aff{Apple}
  \end{tabular}
  \hspace{1em}
  \begin{tabular}[t]{c}
    Melanie Weber \\
    \aff{Harvard University}
  \end{tabular}
  \hspace{1em}
  \begin{tabular}[t]{c}
    Jonathan A.\ Kelner \\
    \aff{MIT}
  \end{tabular}
  \\[2em]
  \begin{tabular}[t]{c}
    Peter L.\ Bartlett \\
    \aff{UC Berkeley \&} \\
    \aff{Google DeepMind}
  \end{tabular}
}

\date{}

\begin{document}

\maketitle

\begin{abstract}
We show that two-layer transformers trained to near-zero loss on a sorting task converge to one of two mechanistically distinct solution classes depending on their initialization. One solution class consists of solutions that do not use the second attention layer to sort, but also fail to length-generalize. The other, which uses both attention layers, length-generalizes perfectly, despite only being trained to sort sequences of a fixed length. By carefully choosing the weight initialization, we demonstrate that we can steer the model toward learning one solution or the other. We identify input types that are rare at the training length but increasingly common at longer sequence lengths as a source of this disparity in length-generalization, and find that layer-wise specialization of the second class of solutions allows them to handle these rare sequences more accurately. In the second class, we discover a mechanism by which the second layer attention can use the contextual information gathered by the first layer attention to reduce its search space. On the related task of finding the minimum of a sequence, we find a similar two solution-type phenomenon, but with the opposite generalization pattern, where a different form of layer-wise specialization causes the two-layer solution to length-generalize worse than the single-attention solution.
\end{abstract}

\section{Introduction}

\begin{figure}[!t]
  \centering
\includegraphics[width=\linewidth]{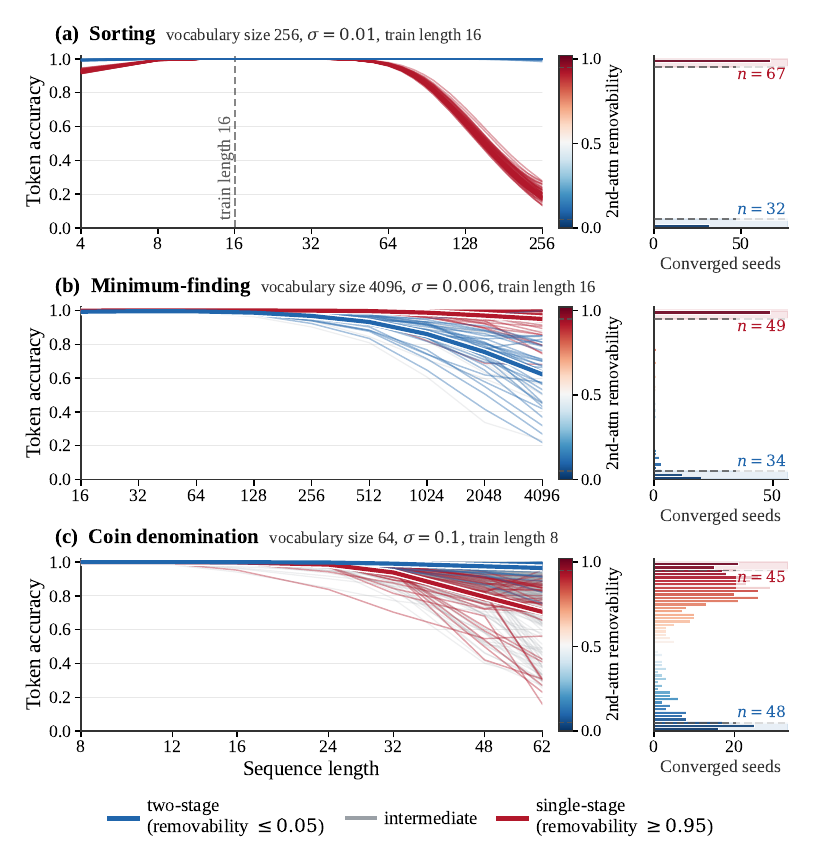}
  \caption{\textbf{Two-layer transformers trained from different initializations learn distinct solution types on three tasks that one layer could solve, and which type length-generalizes depends on the task.} Each curve is one seed's token accuracy, evaluated beyond its training length. Curves are colored by removability, defined as the model's sequence accuracy when the second attention layer is ablated. The side histograms show how removability is distributed across seeds. \textcolor{red}{Single-stage} models are unaffected by the ablation, while \textcolor{blue}{two-stage} models have poor accuracy without the second attention layer. \textbf{(a)}~ Sorting: removability is either 0 or 1, and \textcolor{blue}{two-stage} solutions length-generalize almost perfectly while \textcolor{red}{single-stage} solutions do not. \textbf{(b)}~Minimum-finding: The pattern reverses where \textcolor{red}{single-stage} solutions generalize better than \textcolor{blue}{two-stage} solutions. \textbf{(c)}~Greedy coin denomination: the same pattern as sorting occurs but with a broader spread across seeds and a weaker correlation between solution type and generalization.}
  \label{fig:three-tasks}
\end{figure}

A recurring hope in mechanistic interpretability is that neural networks trained on well-understood tasks will rediscover familiar algorithms, and that similar mechanisms will recur across independently trained models \citep{olah2020zoom, akyurek2023learningalgorithmincontextlearning}. Recent work has complicated this picture by demonstrating settings in which neural networks can implement one of multiple solutions to the same task. For modular addition, \citet{zhong2023clock} identified two mechanistically different solutions, termed the ``Clock'' algorithm and the ``Pizza'' algorithm \citep[see also][]{nanda2023progressmeasuresgrokkingmechanistic,yip2024modular}, and showed that changes in architectural hyperparameters can determine which solution emerges. Multiple solutions with different generalization behavior out of distribution have since been found in other settings, including in-context learning \citep{park2024competition, wurgaft2025context}, state tracking \citep{li2025howlanguagemodelstrack}, and algorithmic tasks such as counting and addition \citep{zhao2025random}. Our paper introduces a new general setting for studying when neural networks converge to one of two qualitatively different solutions. In particular, we study tasks that are possible for a one-layer transformer to implement, and train them with a two-layer transformer.

While common intuition about training dynamics suggests that neural networks have a "simplicity bias" where they tend to converge to simpler solutions, and that simpler solutions generalize better than complex ones~\citep{pmlr-v235-gatmiry24a,du2024hierarchicalsimplicitybias,shah2020pitfalls}, recent work refutes this conventional wisdom~\citep{pmlr-v235-tsoy24a}. We find that under some parameter initializations, these two layer transformers converge to single-stage solutions, where the second attention can be ablated without impacting accuracy, while for other initializations, the transformers converge to a solution that meaningfully uses both attention layers. We find that for the task of \emph{minimum-finding}, the simpler single-layer solutions generalize better. However, for the task of \emph{sorting numbers}, we find that the more complex two-layer solutions generalize better, counter to common intuition about simplicity bias. In a third, \emph{greedy coin denomination} task, we find a larger variation in types of solutions learned, and more mixed results about generalization, but with a correlation between reliance on the 2nd attention layer and length-generalization abilities.

This paper focuses on the task of sorting numbers, where we find the cleanest split in solution types across seeds. We study this task in more detail with mechanistic interpretability, aiming to explain the unintuitive phenomenon of the more complex 2-layer solution length-generalizing better. Sorting a sequence of numbers is a task in $\mathbf{FTC^{\mathrm{0}}}$ which can theoretically be performed with just a single layer transformer given a large enough embedding size. In fact, \citet{weiss2021thinking} show that training a single layer transformer for sorting reaches accuracy close to 1 on the lengths trained on.  

We argue for two mechanisms of why length-generalization occurs in the more complex solution. First, the more complex solution learns a form of cross-layer specialization to more accurately sort sequences with numbers close to each other. This is a property of sequences that is rare at the length trained at, but increasingly common at longer evaluation lengths. Second, we show evidence of a hinting mechanism, where the first layer can provide a range of possible next tokens to the second, which may make it easier to select the correct next token out of a candidate set of potential next tokens that grows with larger input lengths. We additionally show how to steer the model toward each solution type by varying the initialization scale across layers.

\textbf{Organization.} Section~\ref{sec:setup} introduces the task and experimental setup.
Section~\ref{sec:two-algorithms} presents the two solution types and describes each at a high level.
Section~\ref{sec:mechanistic} gives a mechanistic analysis of the two-stage solution, identifying two properties that account for its length-generalization advantage. Section~\ref{sec:steering} investigates what determines which algorithm emerges from training. Section~\ref{sec:alternate_algorithms} discusses results for algorithms other than sorting, and Section~\ref{sec:related} discusses related work.

\section{Setup of the Sorting Task} \label{sec:setup}

\textbf{Task.} The input to the model is a sequence of $K$ integers drawn uniformly at random (without replacement) from a vocabulary $\{1, \dots, N\}$, followed by a separator token \texttt{SEP}, after which the model autoregressively outputs them from smallest to largest. Writing the full sequence as $x_1, \dots, x_K, \text{SEP}, x_{s(1)}, \dots, x_{s(K)}$, where $s$ sorts the input, the query $x_{s(i)}$ must predict $x_{s(i+1)}$ (and $\text{SEP}$ must predict $x_{s(1)}$). We call $g_i := x_{s(i+1)} - x_{s(i)}$ the gap at that position. In most experiments, we train on sequences of length $K = 16$ and evaluate length generalization at every $K$ from $2$ to $N$, the largest possible sequence. 

\textbf{Architecture.} We use a two-layer, causally-masked transformer in the style of nanoGPT, simplified to one attention head per layer to facilitate mechanistic analysis. Crucially, we omit positional encodings \citep{kazemnejad2023impact}, which is a necessary condition for length-generalization on algorithmic tasks as any model that encodes absolute position is structurally prevented from generalizing to sequences longer than those seen during training. Furthermore, in our preliminary experiments, we observed that positional embeddings induce more complicated difficult-to-interpret algorithms, with each token attending to itself in a manner that nonetheless becomes mechanistically necessary for accurate model performance. Studies of multiple heads per layer and deeper transformers can be found in Appendix \ref{app:4-layer-section}.

\textbf{Training.} We train online, sampling a new batch of sequences at each iteration. Due to the large space of possible sequences, it is very unlikely to revisit the same input twice, making the distinction between training data and test data on the same sequence lengths not meaningful. We use the AdamW optimizer with a learning rate of $0.03$ and a batch size of $4096$, optimizing cross-entropy loss over the output tokens. Models reach $100\%$ training accuracy early in training; we then continue training for a total of $60{,}000$ iterations to near-zero loss.

\section{Two Sorting Algorithms Emerge from Initialization}
\label{sec:two-algorithms}

Across different weight initializations, with the training data ordering held fixed, we find that trained two-layer transformers achieving perfect sorting accuracy do not all learn the same algorithm. Figure \ref{fig:three-tasks} shows that when the second attention layer is removed from the trained model, per-seed accuracy splits into two tight clusters. One cluster retains full sample accuracy, while another cluster of seeds has accuracy tightly clustered around 0. This is in contrast to the first attention layer, which can't be removed for any seed without dropping the sample accuracy to 0. The fact that the sample accuracy after ablating the second attention is approximately a Bernoulli random variable suggests that there are at least two qualitatively different algorithms that emerge from training, depending on the weight initialization.

We call the algorithm type that does not depend on the second attention a \textbf{single-stage solution}, and the algorithm type that does depend on the second attention a \textbf{two-stage solution}. The distinction between these two solution types has a striking consequence for length-generalization, as two-stage solutions generalize to sequences longer than those seen during training, while single-stage solutions do not (Figure~\ref{fig:three-tasks}). The remainder of this section describes the single-stage solution at a high level. Section~\ref{sec:mechanistic} then describes how the two-stage solution is different and provides a mechanistic account of why the two-stage solution length-generalizes.

\subsection{Single-stage solutions}

\begin{figure}[!t]
    \centering
    \includegraphics[width=\linewidth]{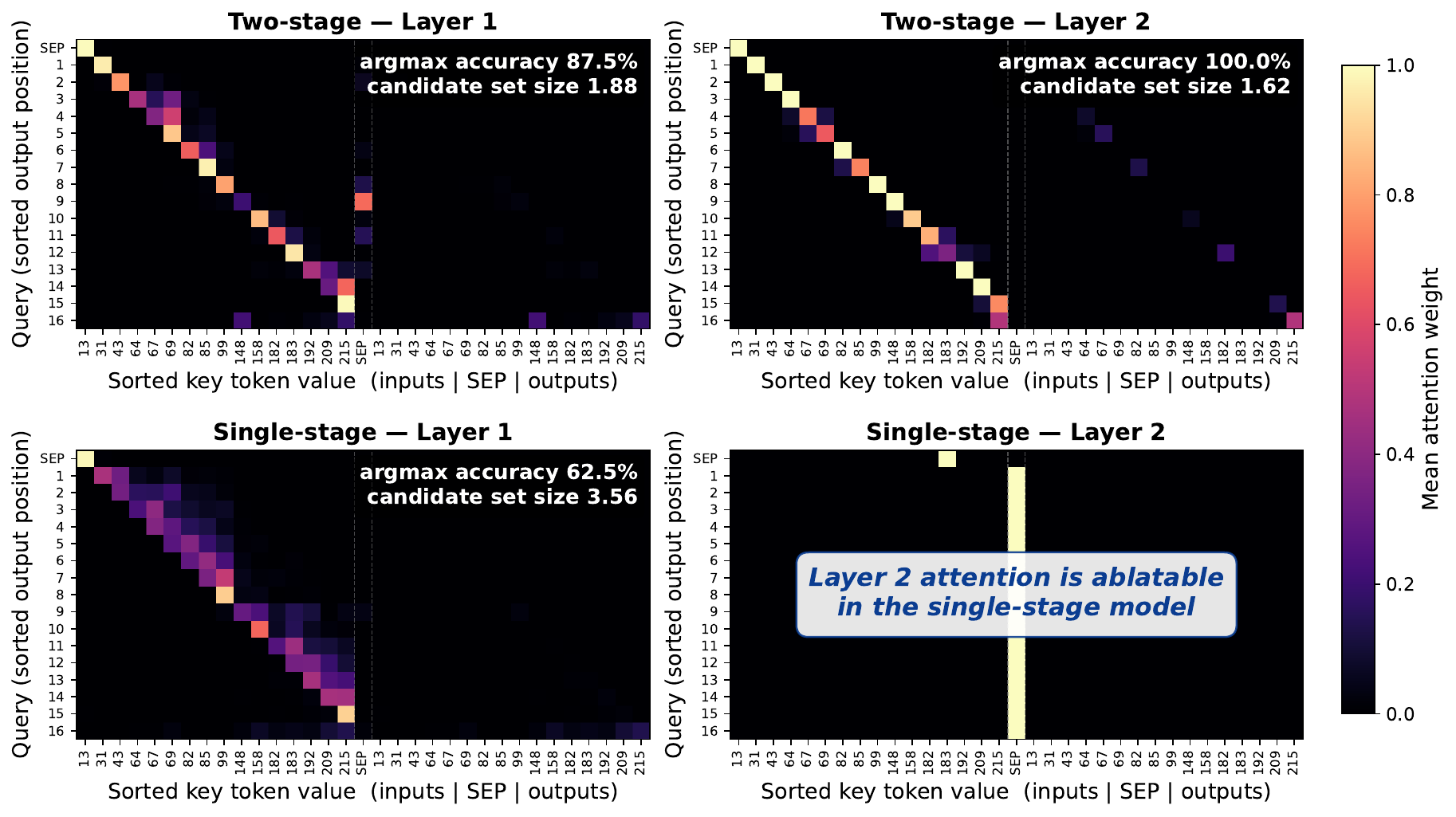}
    \caption{Attention probability heatmaps for a two-stage seed (top) and a single-stage seed (bottom), with tokens ordered by numerical value. Both layers of the two-stage model more reliably attend to the correct next token, whereas the single-stage model's first layer distributes attention across a broader set of nearby candidate tokens.}
\label{fig:ridgeline}
\end{figure}

In single-stage solutions, there is only one attention layer that becomes involved in the computation, and therefore only one query per position to consider. We find that the query at position $x_{s(i)}$ attends most strongly to the position holding $x_{s(i+1)}$ in the unsorted part of the sequence, but also attends partially to positions holding larger values (Figure~\ref{fig:ridgeline}, bottom). The resulting value vector is therefore a linear combination of nearby tokens in sorted order, typically concentrated on the correct one. However, in instances where token attended to most is not the correct next token, the transformer still predicts a correct next token. We hypothesize that the role of the first attention in the single-stage model is to select a set of potential next tokens, which we term the candidate set, consisting of those assigned non-trivial attention probability by the attention. We further hypothesize that the two MLP layers and the unembedding matrix could be thought of as selecting among this candidate set of tokens to produce the correct output.

\section{Two-stage solutions and their Mechanistic Analysis}
\label{sec:mechanistic}
%\subsection{Two-stage solutions.}

We identify two properties of the two-stage solution that may account for its superior length-generalization. First, the two attention layers specialize across gap sizes between consecutive tokens, where the first layer handles tokens with small gaps to the next largest number, while the second layer has a larger role for tokens where the gap to the next largest token is larger (\S\ref{sec:specialization}). Importantly, sequences with small gap sizes lie in the low-probability tail of the training distribution at small training lengths, so a model can achieve high overall accuracy without handling them reliably. However, as sequence length increases, the gap distribution shifts toward smaller values, making this previously rare statistic increasingly likely. Correctly handling small gaps therefore becomes increasingly important for length-generalization. Second, we provide evidence that the first attention layer can supply the second with contextual information that narrows the search for the correct next token to a smaller interval (\S\ref{sec:attn2-qk-mechanism}). This \emph{windowed monotonicity} mechanism may be another factor that allows the second attention layer to maintain an accurate attention distribution even as sequence length grows and tokens become more densely packed.

\subsection{Handling Rare Sequences With Specialization Across Layers}
\label{sec:specialization}

While single-stage solutions to the sorting task are measured to have perfect accuracy, we find that they still have adversarial examples. For example, input sequences that are permutations of consecutive integers, with gaps of $1$ between sorted input tokens, are unable to be sorted by single-stage models, but can be sorted by two-stage models (see Appendix \ref{app:varying_k} for a further study of adversarial examples). As sequence length grows, sequences with smaller gap sizes between tokens become increasingly likely. As we reach an input length equal to the vocabulary size, every gap becomes exactly 1, which we know single stage models are unable to sort, even at short sequence lengths. We hypothesize that the high accuracy of two-stage models on these rare sequences is enabled by a form of layerwise specialization, where tokens of small gaps (whose next largest token is close) are generally handled mostly by the first layer, while predicting large token gaps is handled more by the second layer's attention. We present three pieces of evidence supporting this layerwise specialization hypothesis in this subsection.

\begin{figure}[!t]
  \centering

  \begin{subfigure}[b]{0.48\linewidth}
    \centering
    \includegraphics[width=\linewidth]
      {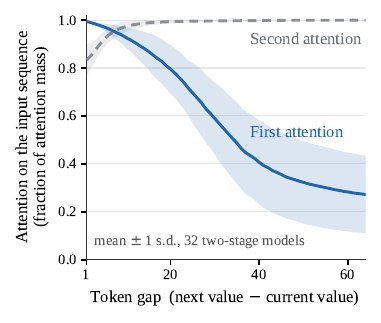}
    \phantomcaption
    \label{fig:specialization_across_layers}
  \end{subfigure}
  \hfill
  \begin{subfigure}[b]{0.48\linewidth}
    \centering
    \includegraphics[width=\linewidth]
      {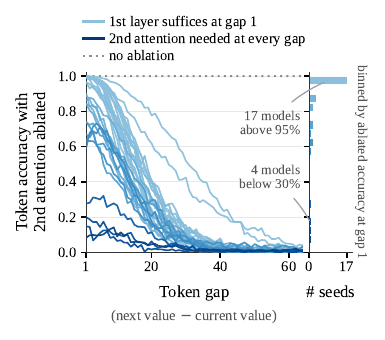}
    \phantomcaption
    \label{fig:removal_of_l2_by_gap}
  \end{subfigure}

  \caption{\textbf{(a)}~Fraction of the first attention's mass on input-sequence tokens as a
    function of gap size, averaged over two-stage models. As the gap increases,
    the first attention increasingly attends to the separator token or sorted-output tokens
    rather than unsorted input tokens, indicating that the first attention may be playing a less significant role in the computation.
    \textbf{(b)}~Token accuracy after removing the second attention, broken down by gap size,
    across two-stage seeds. For many two-stage seeds, even though the accuracy on random inputs goes to 0 after removing the 2nd attention, for tokens of gap size 1, the accuracy remains at $100\%$ after the ablation.}
  \label{fig:gap-specialization}
\end{figure}

\paragraph{The first attention provides weaker contextual information at large gaps, while the first layer often suffices at small gaps.}
As the gap increases, the first attention increasingly attends less to the input sequence, attending more to the separator token, which may act as an attention sink, and also to already sorted inputs, which cannot contain the correct next largest value (Figure~\ref{fig:specialization_across_layers}). This suggests that the first layer's contribution diminishes as the gap size grows. While for all single-stage seeds, removing the second attention leaves accuracy at $100\%$, at gap size $1$, most two-stage seeds also have accuracy remain at $100\%$ when the second attention is removed (Figure~\ref{fig:removal_of_l2_by_gap}). This suggests that any ``two-stage'' computation matters mainly at medium to large gaps. A minority of seeds, however, do lose accuracy even at gap~$1$. This diversity suggests that multiple valid solutions exist within the two-stage family, differing in the extent to which the two layers share responsibility at small gaps. 

\begin{figure}[!t]
  \centering
  \includegraphics[width=\linewidth]%{figures/hijack_allI_avg_seeds_fl.png}
{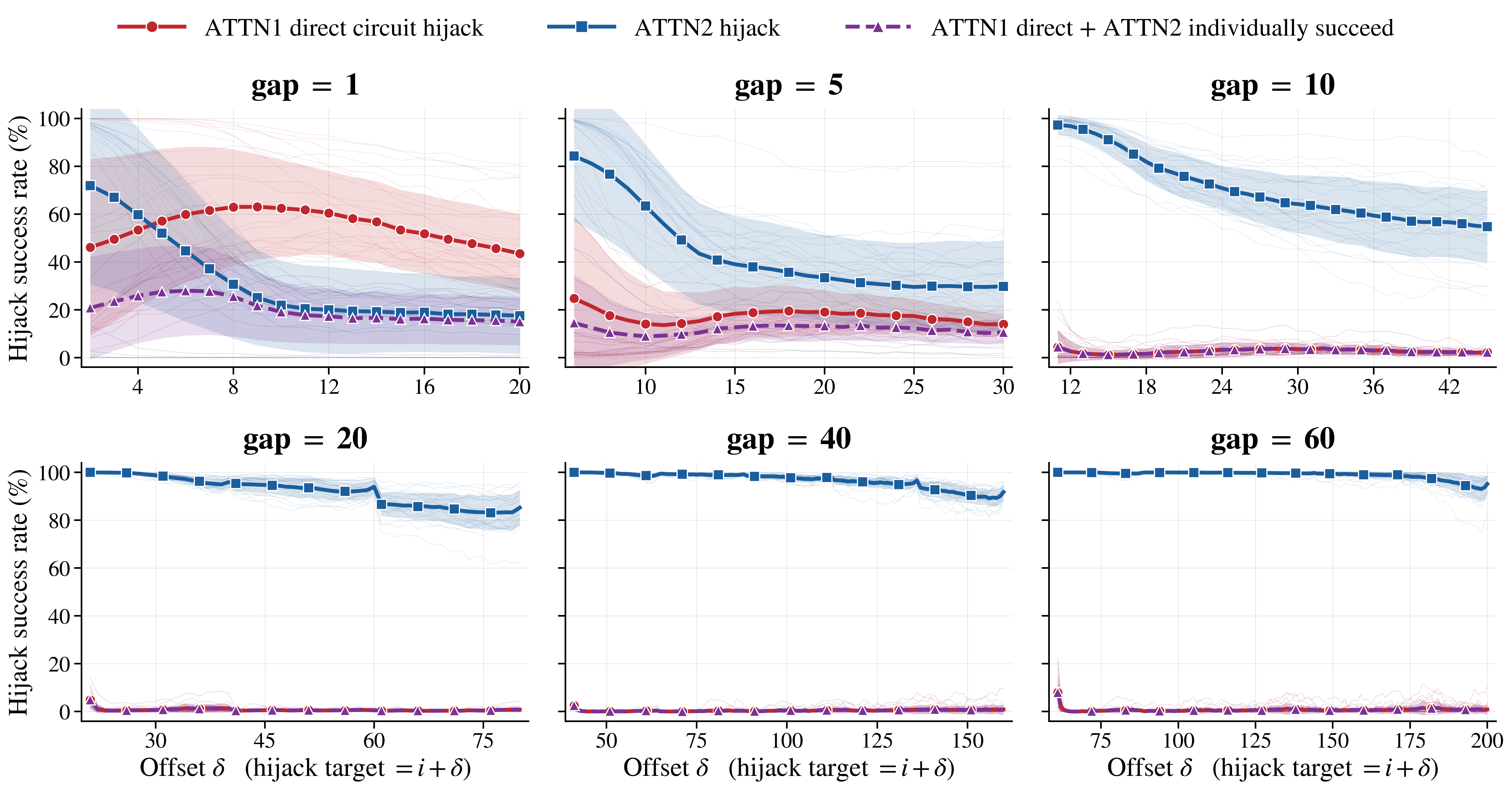}
  \caption{Hijack success rates averaged over all values of~$i$ and 23~two-stage
    seeds (sequence length $32$, vocabulary size $512$). Each panel shows a different gap~$g$. Red:
    redirecting the first attention's direct circuit (bypassing the second attention). Blue: redirecting
    the second attention's focus. Purple (dashed): fraction of instances where both
    interventions independently succeed on the same input. At small gaps, both
    circuits contribute, to varying extents depending on the seed; as the gap grows, the second attention dominates entirely. 
    % \nathan{As this is the only experiment with vocab size 512, I am concerned that readers will perceive this as cherry-picking.}
    }
  \label{fig:hijack-fl-avg}
\end{figure}

\paragraph{The first layer's direct path loses causal control at large gaps.}
We corroborate this division of labor across gap sizes causally with two interventions. First, we force the first attention to attend to a chosen token only along the pathway that bypasses the second attention, leaving its input to the second attention unchanged.  Second, we force the second attention to attend to a chosen token. For each, the \emph{hijack success rate} is the fraction of interventions that change the prediction to the chosen token. Figure~\ref{fig:hijack-fl-avg} reports the mean hijack success rates of both interventions across six gap sizes, averaged over 23~two-stage seeds (sequence length~$32$, vocabulary size~$512$). At the smallest gap, there is a very large spread across seeds, similar to the spread seen in Figure~\ref{fig:removal_of_l2_by_gap}. The co-success rate at small gaps is approximately 20\%, indicating that both pathways can influence the model's prediction. As the gap grows, the first attention's direct circuit loses influence, while the second attention hijack success rate remains close to $100\%$ across all gaps, corroborating results found in Figure~\ref{fig:specialization_across_layers}.

\subsection{How the first attention can assist the second}
\label{sec:attn2-qk-mechanism}

While at very small or very large gaps the two-stage model primarily specializes across layers, we believe that at intermediate gaps, common at the training length or medium-long context, both layers play a compositional role in predicting the next token. In this section we describe how we believe the first layer can provide contextual input to the second and how this may assist with length generalization at intermediate lengths.

At average gap sizes, in two-stage models, the second attention markedly attends more reliably to the correct next token than the first, namely ${\approx}99.5\%$ of the time, versus ${\approx}86\%$ for the first (Figure~\ref{fig:ridgeline}). We hypothesize that this difference in accuracy is explained with a hinting mechanism where attention layer 1 supplies attention layer 2 with contextual information that effectively provides an \emph{upper bound} on the next correct number. This hint originates from the query side, where the first attention gathers contextual information about the current output position and passes it forward through the residual stream, giving the second attention an estimate of where the next sorted value lies. The second attention may then use this signal to adjust the current token's representation so that its attention is concentrated on the narrow range between the current value and the upper bound provided by the first attention, rather than searching over the entire input sequence.

\paragraph{Isolating the effect of contextual information gathered by the first attention on the second attention.} To better understand how the contextual information gathered by the first attention affects the second attention score, we consider the stylized case where the first attention is concentrated on a single token in the context. We can then analytically characterize the second attention score as a function of four quantities: the query token~$z$, the key token~$t$, the first attention target on the query side~$x$, and the first attention's target on the key side~$y$:
\begin{align}
  s(z, x, t, y)
  \;=\;
  \mathbf{q}(z, x)^\top \mathbf{k}(t, y),\label{eq:attncontextual}
\end{align}
where $\mathbf{q}(z,x) = W_Q^{(2)} \operatorname{LN}\!\bigl(
\operatorname{MLP}_1\!\bigl(\operatorname{LN}(e_z + V^{(1)}_x)\bigr)\bigr)$
and $\mathbf{k}(t,y)$ is defined analogously. Here $V^{(1)}_x$ denotes the
first-layer attention's value vector for token~$x$ (projected through $W_O^{(1)}$).
The reason we only look into the MLP part of the residual stream after the first layer in Equation~\ref{eq:attncontextual} is because, as we show in Appendix~\ref{app:attn2-mlp1-dependence}, the second attention's QK scores depend almost exclusively on the first MLP's output rather than the full residual stream.

%%%
Interestingly, as we visualize this score (Appendix~\ref{app:attn2-mlp1-dependence}), it is essentially invariant to the key-side target~$y$ of the first attention, but strongly modulated by its query-side target~$x$, with a high-score band emerging in the interval $[z, x]$. That is,
the second attention uses contextual information from the first attention exclusively on the query side, consistent with the hinting mechanism described above.

\begin{figure}[htbp]
  \centering
  \begin{subfigure}[b]{0.48\linewidth}
    \centering
    \includegraphics[width=\linewidth]{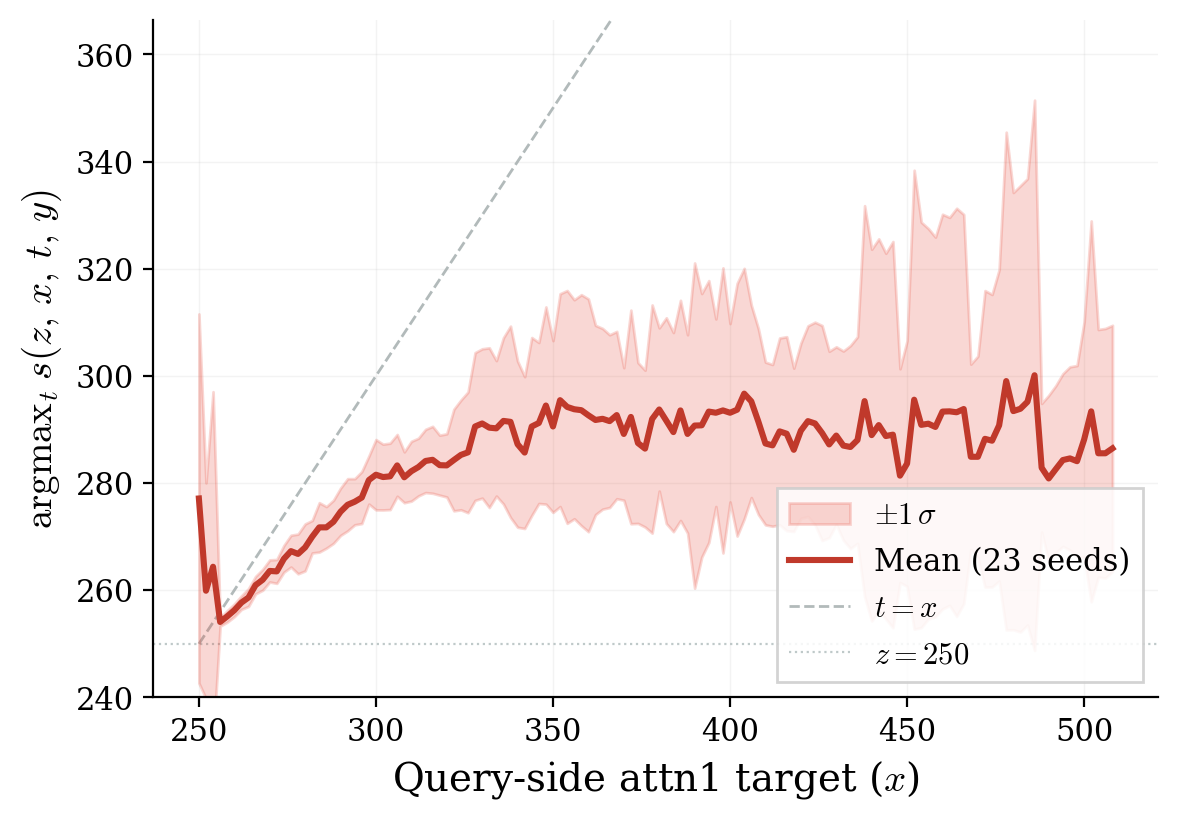}
    \phantomcaption
    \label{fig:argmax-saturation}
  \end{subfigure}
  \hfill
  \begin{subfigure}[b]{0.48\linewidth}
    \centering
    \includegraphics[width=\linewidth]{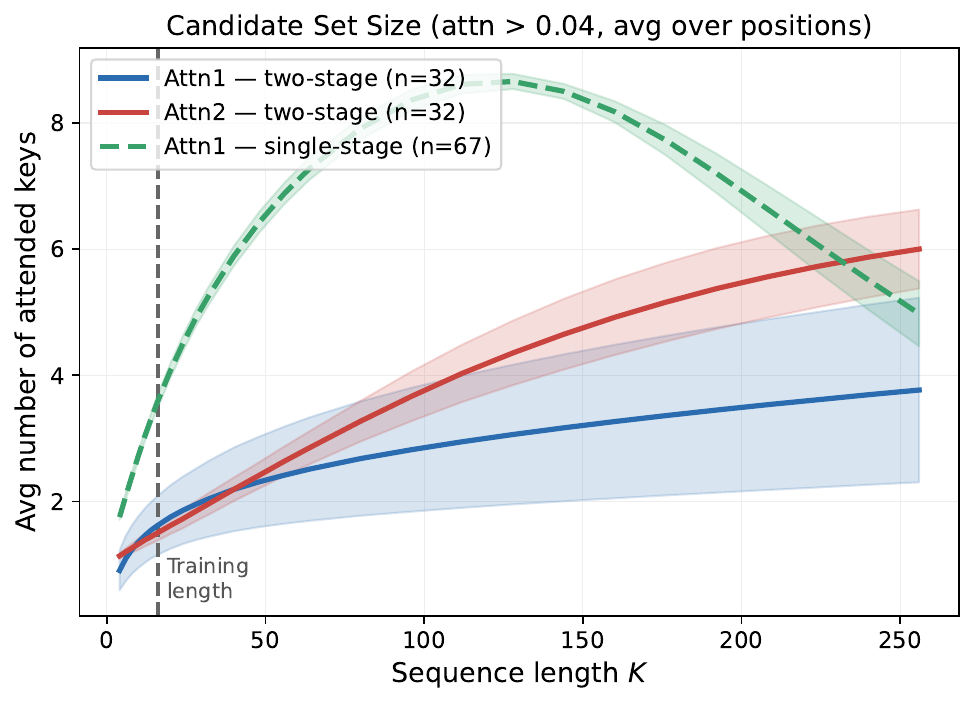}
    \phantomcaption
    \label{fig:attn-nkeys-vs-seqlen}
  \end{subfigure}

  \caption{\textbf{(a)} Most-attended key of the second attention as a function of the query-side token $x$ that the first attention is forced to attend to. The second attention's focus typically lies between the current token and the first attention's target, increases approximately monotonically, and saturates for large first attention targets. For small interventions, resembling when the next token is very close in value to the current token, occurring at large input lengths, this property does not hold. \textbf{(b)}~Average number of keys with attention weight ${>}0.04$ for varying sequence length for a two-stage and single-stage model.}
  \label{fig:saturation-and-nkeys}
\end{figure}

\textbf{Windowed monotonicity.}
Given an upper-bound hint~$x$ from the first attention, the second attention does not need to learn a globally correct ranking over the entire vocabulary; it suffices to correctly classify tokens within the window $[z, x]$. We hypothesize that this reduced search problem contributes to the second attention's higher accuracy relative to the first.  Figure~\ref{fig:argmax-saturation} is consistent with this picture. We intervene on
which token the first attention attends to on the query side and measure which key token the second attention attends to most strongly, for a current token of value $250$ (the choice is arbitrary), averaged over 23~seeds. The second attention's most-attended key usually lies between $z$ and the first attention's target, rarely exceeds that target, and increases approximately monotonically with it.

When the first attention's target is pushed far above the current token, the curve saturates, with the second attention's most-attended key leveling off at roughly $300$ regardless of how large the target becomes. One interpretation is that the second attention limits how far it follows the upper-bound hint, and stays closer to the local neighborhood of $z$ when the hint lies far above it (a situation likely very rare during training). Within their respective windows, the second attention's QK scores also appear smoother than the
first attention's scores
(Appendix~\ref{app:smoothness}), consistent with its narrower search range. The analysis above assumes the first attention is fully concentrated on a single token. When we instead split its weight equally among $n \in \{2,3,4\}$ tokens spaced $\delta$ apart, the band structure persists, and the argmax stays biased toward the smallest token in the attended set rather than the mean (Appendix~\ref{app:split-attn1}), suggesting the mechanism is robust to distributed first-layer attention.

\textbf{Smaller candidate sets.}
We measure the number of keys receiving attention weight above $0.04$, averaged over output positions, as sequence length grows (Figure~\ref{fig:attn-nkeys-vs-seqlen}). At the training length, both attention layers of two-stage models attend to about $1.5$ keys, compared to about $3.6$ for the first attention of single-stage models. Beyond the training length, the first attention of two-stage models keeps a smaller candidate set than its single-stage counterpart at every length we test (i.e. $3.1$ versus $8.7$ keys at $8$ times the training length). Since small gaps become more common as sequences grow and the first attention primarily handles small gaps in two-stage models (\S\ref{sec:specialization}), this is the layer we expect to matter most at long lengths. The second attention's candidate set grows faster, reaching about $6$ keys at $16$ times the training length. This is in particular aligned with our observation on length-generalizability of single-stage vs two-stage models in Figure~\ref{fig:three-tasks}.

\section{What Determines Which Algorithm Emerges?}\label{sec:steering}
\subsection{Steering via Initialization Scale}
 The results so far show that at least two mechanistically distinct solutions coexist in the same loss landscape, with different length-generalization behavior, and the initialization seed determining which one is converged to. If the goal is a model that length-generalizes, one must either train many models and select the generalizing one post hoc, or find a way to bias training toward the desired solution. We show that adjusting the relative initialization scale between the two layers is sufficient to do the latter, enabling the steering of nearly all models towards a pre-specified solution type.

\begin{figure}[htbp]
    \centering
    \includegraphics[width=\linewidth]{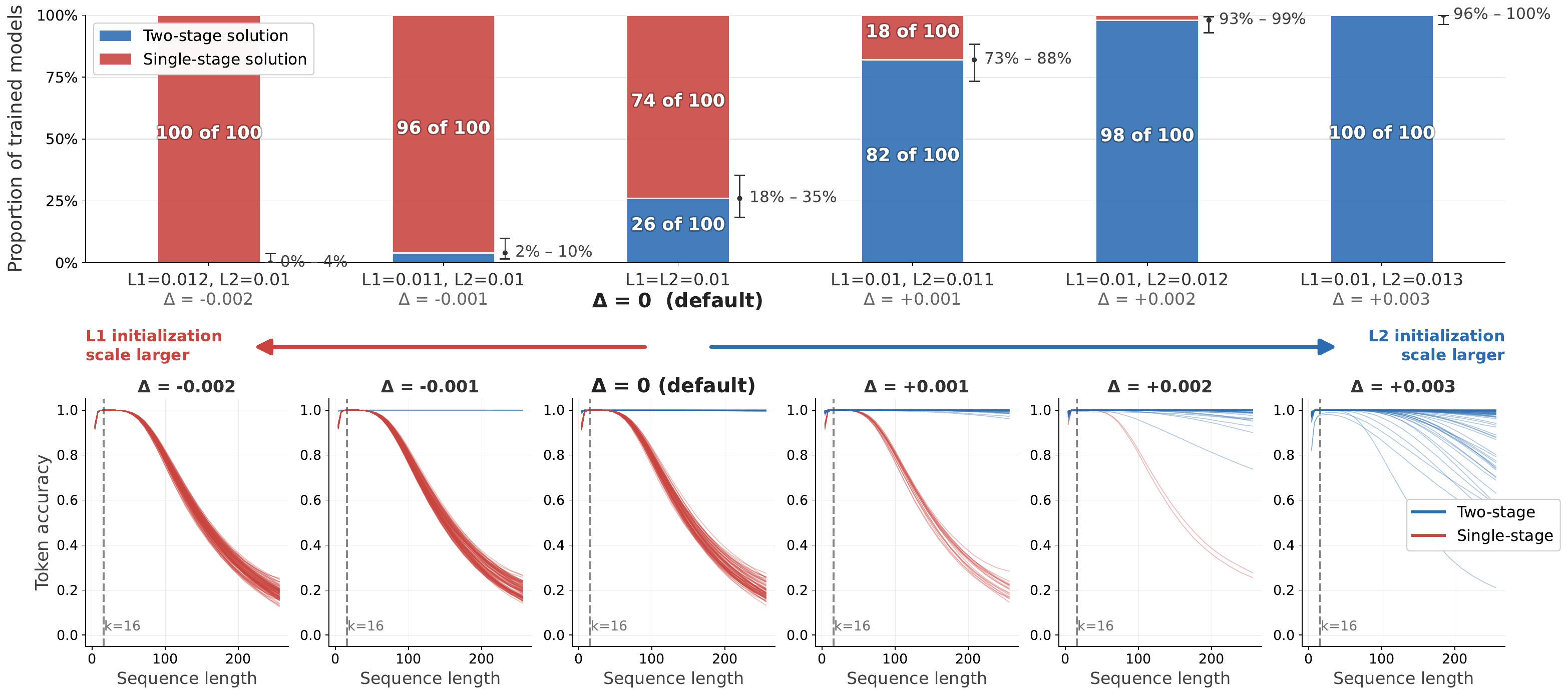}
    \caption{\textbf{Top:} Fraction of seeds converging to the two-stage solution as a function of the relative initialization scale difference $\Delta = \sigma_2 - \sigma_1$ (sequence length 16, vocabulary size 128, 100 seeds). A small positive $\Delta$ drives all seeds to the two-stage solution and a small negative $\Delta$ drives all seeds to the single-stage solution. \textbf{Bottom:} Length-generalization curves for all models in each condition, colored by solution type. \textcolor{blue}{Two-stage} models generalize beyond the training length, while \textcolor{red}{single-stage} models do not.}
    \label{fig:steering_figure}
\end{figure}

We fix one layer's weight initialization scale at $\sigma=0.01$ and increase the other's to $0.01+|\Delta|$ (where $\Delta > 0$ makes layer 2 larger while $\Delta < 0$ makes layer 1 larger), training 100 initialization seeds per condition. At the symmetric baseline, 26\% of models are two-stage, while increasing Layer~2's initialization scale to just $\sigma_2 = 0.013$ drives this to 100\% (Figure~\ref{fig:steering_figure}). Conversely, increasing Layer~1's scale to $\sigma_1 = 0.012$ drives it to 0\% (Figure~\ref{fig:steering_figure}). A plausible explanation is that layers initialized to be larger develop structure first during training as a larger initialization scale makes initial gradients larger. When the first layer is initialized larger, it may learn to solve the task before the second layer ever gets involved, which is consistent with the observation that in single-stage models, the second attention layer almost never attends to the token that the model predicts as the next token, suggesting that it may never have been under gradient pressure to become useful. Models steered this way often retain their solution type's generalization behavior, with the exception of a small cost when the 2nd layer's scale is pushed too far, discussed further in Appendix \ref{app:steering_length_gen}. For the effect of absolute initialization scale rather than relative scale, see Appendix \ref{app:symm_init_scale}.

\subsection{Task difficulty and model capacity bias which solution emerges}
\label{sec:task-difficulty} \label{sec:model-capacity}
Which solution emerges also depends on hyperparameters controlling the task difficulty and the model expressivity. Increasing the vocabulary size $N$ favors the two-stage solution. In particular, nearly all seeds are single-stage at $N=64$, while most are two-stage at $N=512$ (Figure \ref{fig:varN}), which we hypothesize is because consecutive values become harder for a single attention layer to tell apart (see Appendix \ref{app:vocab_exp}). Increasing the embedding dimension $E$ at fixed $N=256$ has the opposite effect (Appendix~\ref{app:embed_exp}) where the portion of two-stage seeds falls from every seed at $E=32$ to none at $E=80$, possibly because wider models can distinguish tokens with a smaller weight norm, which is easier to reach from a small initialization. Together, these suggest that the second layer is recruited when the task is hard relative to the capacity available to solve it (see Appendix \ref{app:embed_exp} for details). Along with task difficulty and model capacity, we study other factors impacting which solution emerges in the appendix, including the training length, and find a non-monotonic relationship with the solution type (Appendix \ref{app:varying_k}).

\section{Do these phenomena persist across other algorithms?}
\label{sec:alternate_algorithms}

While this paper has focused on sorting, where the split between solution types is cleanest, the phenomenon is not specific to sorting. We also study two related tasks that are possible to implement with a single transformer layer, but train them using the same two-layer architecture. Across these tasks, we again find that different initialization seeds lead to solutions that differ in the extent to which they use the second attention layer. We also find that the solution type predicts length-generalization. However, the relationship between using the second layer and length-generalization is not the same across tasks.

The three tasks we study are closely related conceptually. Sorting repeatedly finds the next larger element in the input, while minimum-finding performs only the first step of this procedure, and greedy coin denomination repeatedly finds the largest denomination that does not exceed the current remainder and deducts it from the remainder. Thus, sorting and coin denomination both involve successor or predecessor queries, which only require one layer of a transformer to perform. Despite these similarities, the three tasks produce different distributions of solution types and different patterns of length-generalization.

\paragraph{Minimum-finding.}
We train the same two-layer architecture to find the minimum of a length-$K$ sequence of integers drawn from $\{1,\ldots,N\}$ with replacement. The input consists of the sequence followed by a separator token, and the model outputs the minimum value. We train at $K=16$ with $N=4096$ and evaluate at longer sequence lengths.

One important distinction between the sorting task and the minimum-finding task is that for the minimum-finding task, we sample input sequences with replacement. Sorting with duplicates cannot be performed by a single attention layer without positional encodings, since the model must keep track of how many copies of each value to output. Minimum-finding, in contrast, remains possible with one layer because the output does not depend on a value's multiplicity. Minimum-finding therefore lets us compare single-stage and two-stage solutions on inputs with duplicates.

\paragraph{Greedy coin denomination.}
We also study the task of greedy coin denomination. The input is a set of distinct denominations, followed by a separator token and a target value. The model must output the sequence of remainders produced by repeatedly subtracting the largest denomination that does not exceed the current remainder. We train on denomination sets of size $K=8$ and evaluate on larger sets.
This task is closely related to sorting because at each output step, the model must find the largest available denomination below the current remainder. This is a predecessor query, whereas sorting requires a successor query. We study the greedy algorithm rather than the optimal change-making algorithm since it is an easier task that can be done with a single-layer transformer, although it is not optimal for every set of denominations.

\paragraph{Choice of vocabulary and initialization scale.}
For both tasks, we choose the vocabulary size so that it is small enough for a single-stage solution to be learnable, but large enough that the model sometimes learns to use the second layer. Since minimum-finding is easier than sorting, we use a larger vocabulary for minimum-finding ($N=4096$, compared with $N=256$ for sorting) before two-stage solutions appear at appreciable rates. We then choose a symmetric initialization scale at which both solution types occur at similar rates. For minimum-finding, we use $\sigma=0.006$, selected from a scan over $\sigma\in[0.005,0.01]$; for coin denomination, we use $\sigma=0.1$.

\subsection{Different tasks produce different solution types}
Figure~\ref{fig:three-tasks} shows the results for all three tasks. As with sorting, ablating the second attention layer separates models according to how much they use the second layer. Some models are almost completely unaffected by the ablation, indicating that they have learned a single-stage solution. Other models fail after the ablation, indicating that they use both attention layers.
For sorting, the two-stage models length-generalize better than the single-stage models. Greedy coin denomination shows the same overall pattern, where models that rely more on the second attention layer tend to generalize better, although the solutions form a more continuous range than they do for sorting.
Minimum-finding shows the opposite pattern, where the single-stage models generalize better than the two-stage models. Whether the two-stage solution aids length-generalization is therefore a property of the task.

\subsection{Competing failure modes in minimum-finding}

The minimum-finding result can be understood by comparing the adversarial examples of the two solution types. Both solution types achieve high accuracy at the training length, but they fail on different types of inputs. In particular, two-stage models degrade when many input values are close to the minimum. These inputs are rare at the training length, but become increasingly common as the sequence length grows. Single-stage models are more robust to these inputs, but are less robust than two-stage models to inputs involving duplicate numbers.

One hypothesis is that duplicates are difficult because they require more precise attention scores. Suppose the minimum has pre-softmax score $a_1$, while each of $n$ copies of the second-smallest value has score $a_2$. To place more total attention on the minimum, the model needs $e^{a_1} > n e^{a_2}$, or equivalently $a_1-a_2 > \log n$.
Thus, the required score margin grows with the number of copies of a competing value. The additional layer in the two-stage models may help make these distinctions. This would explain why they handle some duplicate-related inputs better, but fail on crowded near-minimum inputs.

These results suggest that length-generalization depends on how a solution's failure modes interact with the shift in input distribution. In sorting, the two-stage solution handles small-gap inputs, which become more common at longer lengths. In minimum-finding, the two-stage solution instead appears to fail on inputs with many values close to the minimum, which are the ones that become more common, while performing better on some inputs involving duplicates. More generally, additional expressivity may help a model handle some input types that are rare at the training length, and whether this improves length-generalization depends on whether those input types become important at longer lengths. Studying the minimum-finding and coin denomination solutions in more depth, and building a more general understanding of how and when this occurs, is an interesting direction we leave for future work.

\section{Related Work}
\label{sec:related}

\subsection{Transformer sorting and length-generalization}

Prior work studies sorting as a canonical algorithmic task for transformers and situates it within the broader problem of length-generalization \citep{anil2022exploringlengthgeneralizationlarge}. It has been shown that transformers can be constructed to sort \citep{lindner2023tracr, awasthi2023improving}, and that whether trained ones length-generalize depends strongly on properties of the set-up, such as positional encodings \citep{kazemnejad2023impact}. \citet{yan2020neuralexecutionengineslearning} also found that transformers trained to sort have difficulty distinguishing numerically close inputs, and separately that their attention loses sharpness beyond the training length. We connect the two observations by noting that that close values are rare at the training length but common at longer ones, and show how the two solution types handle them differently through layer-wise specialization. 
More broadly, \citet{zhou2023algorithms} propose the RASP-Generalization Conjecture, which predicts that transformers tend to length-generalize when a task admits a short RASP-L program that works for all input lengths, which includes sorting. Our results suggest a complementary complication, that even when such solutions exist, training need not reliably recover the same one \citep[cf. the seed sensitivity reported by][]{zhou2024transformersachievelengthgeneralization,zhao2025random}, and different learned algorithms can have different length-generalization behavior. While these works help answer the question of whether transformers \emph{can} length-generalize, our work focuses on the analysis of what models learn in practice, studying internal mechanisms of learned algorithms, what drives the model to converge to different solutions, and how to control this process.

\subsection{Mechanistic analysis and multiple learned solutions}

\citet{huang2026discovering} show that length-generalizing transformers often admit compact, interpretable RASP subprograms that suffice for their behavior, providing direct evidence that trained models can internally implement simple algorithmic programs. \citet{urdshals2025structure} study how internal structure develops in a list-sorting transformer during training, identifying emergent head specializations shaped by the data, but don't study the behavior in more complex, multiple layer models, or study length-generalization. \citet{awasthi2023improving} probe a two-layer sorting transformer and find successor-like computation after the second attention layer that the final MLP then cleans up, resembling our two-stage solution. \citet{li2022systematicgeneralizationemergentstructures} also train two-layer transformers on a multi-feature sorting task that their single-layer model fails to learn, and study how the learned solutions decompose the task across layers.

A separate line of work asks whether models trained on the same task converge to a single mechanism or to multiple distinct solutions. \citet{zhong2023clock} argue that even simple tasks can admit qualitatively different mechanistic solutions, while recent follow-up work argues that some of these apparent differences may be different weight/neuron instantiations of the same underlying algorithm, which is reflected in the underlying geometry \citep{moisescu2025geometry}. In in-context learning, models shift between memorizing and generalizing strategies depending on data diversity and training time \citep{park2024competition, wurgaft2025context}. % can say a bit about how our setting is different
% should cite https://arxiv.org/abs/2407.11542
% and should say a little more about the clock and the pizza
Multiple learned mechanisms have also been reported for state tracking \citep{li2025howlanguagemodelstrack}, and for algorithmic tasks such as counting and addition, where outcomes vary substantially across random seeds \citep{zhao2025random}.
% also would be good to cite https://arxiv.org/html/2609.38081v1 

\subsection{Initialization scale and attention dispersion}

\citet{zhang2025initializationcriticaltransformersfit} show that the global initialization scale determines whether transformers learn compositional or memorizing solutions. However, they focus on a different setting, and we instead steer via the relative scale of two layers, where both of our solutions are exact in distribution. \citet{velickovic2025softmaxforsharpsize} show that softmax attention necessarily disperses as the number of keys grows; the two-stage solution's hint mechanism can be seen as one way a model counteracts this, by shrinking the candidate set (\S\ref{sec:mechanistic}).

\section{Conclusion}
\label{sec:conclusion}
In this paper, we studied what transformers do with capacity they do not need. Every task we considered can be solved by one attention layer, yet a two-layer model sometimes leaves the second layer unused and sometimes recruits it, depending on the random initialization, and length generalization abilities vary in these different cases. Whether a model uses capacity it does not need is an outcome of training that we find can be steered, here by the relative initialization scale of the layers. Because the more complex solution generalizes better for some tasks and worse for others, contrary to common intuition, it is important to take away that simplicity alone does not predict out-of-distribution behavior. Importantly, we unravel the mechanism by which the second attention uses the contextual information gathered by the first attention to provide a more accurate prediction. An interesting open question is whether we can characterize such cooperative structures between layers in deeper sorting models. Other open questions include whether the eventual solution can be predicted reliably early in training, and whether the same seed-dependent selection occurs in larger models trained on natural data.

\section*{Acknowledgments}

We thank Cameron S. Allen and Guanghao Ye for helpful discussions early in the project and Stuart J. Russell for feedback. We also thank Shan Chen for assistance in the design and execution of initial experiments.

\section*{Author Contributions}

\textbf{KG} started the project performing an in-depth mechanistic analysis of the single-stage sorting solution. \textbf{NH} found that there was more than one solution type transformers learned depending on the random seed, and \textbf{KG} and \textbf{NH} worked to characterize and analyze the solution types, with specific contributions outlined below.
\textbf{NH} identified the relevance of token gap size, finding adversarial examples in single-stage models and discovering the layer-wise specialization to different gap sizes in two-stage models (Figures \ref{fig:specialization_across_layers}, \ref{fig:removal_of_l2_by_gap}). 
%\textbf{KG} and \textbf{NH} studied what factors affect which solution type emerges (\textbf{NH} Appendix \ref{app:init_scale} and Figure \ref{fig:varN}, \textbf{KG} Figure \ref{fig:varE}). \nathan{KH's proposed rephrasing is missing important nuance in how things happened and also excluding work that I did.}
\textbf{NH} led the study of what factors affect which solution type emerges, discovering how one can control which model emerges based on the initialization (Figure \ref{fig:steering_figure}) and analyzing other factors impacting which solution emerges (Appendix \ref{app:init_scale}, \ref{app:vocab_exp}, and \ref{app:varying_k}). \textbf{KG} later contributed to this exploration studying the impact of embedding sizes in Appendix \ref{app:embed_exp}. \textbf{NH} also discovered a broader set of tasks exhibiting single- and two-stage solutions (Figure \ref{fig:three-tasks} and Section \ref{sec:alternate_algorithms}) and studied how results extend to settings with models deeper than 2 layers and models with more than one head (Appendix \ref{app:4-layer-section}).

\textbf{KG} did extensive experiments to understand how robust single and two-stage models are with respect to manipulating their attention scores. These culminated in \textbf{KG} designing the hijacking experiments (Figure \ref{fig:hijack-fl-avg}), which causally characterized how direct first-layer circuits and first-to-second-layer hinting circuits contribute separately and jointly to the output, extending and corroborating \textbf{NH}'s observations of these distinct computational routes. \textbf{KG} also characterized how the contextual information gathered by the first-layer attention is used by the second-layer attention in two-stage models (Appendix \ref{app:split-attn1},~\ref{app:smoothness}, Section \ref{sec:attn2-qk-mechanism}).
%\khasha{@nathan the previous sentence i think is more accurate because the collaboration of two attentions also includes other study ablations when first attention score is nontrivial on multiple numbers or independence of key side from the hint of first attention} \nathan{To clarify, is the sentence above proposed to take place of the one below these comments?} 
\textbf{KG} discovered a mechanism by which the first layer can provide hints to the second to search in a smaller space, termed the windowed monotonicity mechanism (Figure \ref{fig:argmax-saturation}), providing an explanation for the second layer having more accurate attention probabilities than the first in some settings, initially identified by \textbf{NH}. %\nathan{As I have expressed previously, I feel that saying that it was done independently is inaccurate accounting of our contributions. KG previously wrote "[experiments he did were] after realizing that some seeds the second layer attention is necessary. this I was informed in one of the early joint meetings we had with chloe and shan". While in this quote he claims to have first found out about two-stage models and have done these experiments after this meeting, I have evidence of showing him these experiments on two-stage models prior to this meeting (in fact, they were also shown to him in our very first meeting, prior to Chloe and Shan joining the project). It is really important to me to have accurate credit assignment for the work I initially put into studying properties of two-stage models. Two proposed ways of doing that are either by saying "initially identified by NH" here or mentioning this notion of argmax accuracy across depth separately in the previous paragraph if KG would prefer, for example adding "\textbf{NH} studied properties of two-stage solutions, identifying that the second attention layer can exhibit more accurate attention probabilities in two-stage models" before leading into the specialization clause in the previous paragraph.} 
\textbf{KG} also analyzed the size of the set of keys with nontrivial attention probability and its connection to the length generalization across different solution types (Figure \ref{fig:attn-nkeys-vs-seqlen}). \textbf{KG} also characterized which circuits actively contribute to the residual stream in a typical two-stage model checkpoint (Appendix \ref{app:attn2-mlp1-dependence}, \ref{sec:residual-cleanup}). %the role of MLPs in shaping the input embeddings in the residual stream to be used by the attentions, where more detail can be found in Appendix Sections \ref{app:attn2-mlp1-dependence}–\ref{app:hijacking}.

\textbf{CS} contributed to the length-generalization experiments and played a substantial role in the early experimental design, framing, and presentation of the paper. All authors contributed to the writing of the manuscript.

\section*{AI Use Statement}
In this work, we used generative AI for writing and executing code for our experiments, to generate plots, to help us identify relevant literature, and to assist us with writing. We have reviewed all AI-assisted work. Specifically, all LLM-generated code was verified and tested for correctness by the authors. We take full responsibility for the final content of this work, including the final manuscript and artifacts produced with the aid of generative AI.

\bibliographystyle{plainnat}
\bibliography{paper}

@misc{li2022systematicgeneralizationemergentstructures,
      title={Systematic Generalization and Emergent Structures in Transformers Trained on Structured Tasks}, 
      author={Yuxuan Li and James L. McClelland},
      year={2022},
      eprint={2210.00400},
      archivePrefix={arXiv},
      primaryClass={cs.LG},
      url={https://arxiv.org/abs/2210.00400}, 
}

@misc{yan2020neuralexecutionengineslearning,
      title={Neural Execution Engines: Learning to Execute Subroutines}, 
      author={Yujun Yan and Kevin Swersky and Danai Koutra and Parthasarathy Ranganathan and Milad Hashemi},
      year={2020},
      eprint={2006.08084},
      archivePrefix={arXiv},
      primaryClass={cs.LG},
      url={https://arxiv.org/abs/2006.08084}, 
}

@misc{zhou2024transformersachievelengthgeneralization,
      title={Transformers Can Achieve Length Generalization But Not Robustly}, 
      author={Yongchao Zhou and Uri Alon and Xinyun Chen and Xuezhi Wang and Rishabh Agarwal and Denny Zhou},
      year={2024},
      eprint={2402.09371},
      archivePrefix={arXiv},
      primaryClass={cs.LG},
      url={https://arxiv.org/abs/2402.09371}, 
}

@misc{anil2022exploringlengthgeneralizationlarge,
      title={Exploring Length Generalization in Large Language Models}, 
      author={Cem Anil and Yuhuai Wu and Anders Andreassen and Aitor Lewkowycz and Vedant Misra and Vinay Ramasesh and Ambrose Slone and Guy Gur-Ari and Ethan Dyer and Behnam Neyshabur},
      year={2022},
      eprint={2207.04901},
      archivePrefix={arXiv},
      primaryClass={cs.CL},
      url={https://arxiv.org/abs/2207.04901}, 
}

@misc{velickovic2025softmaxforsharpsize,
      title={Softmax is not Enough (for Sharp Size Generalisation)}, 
      author={Petar Veličković and Christos Perivolaropoulos and Federico Barbero and Razvan Pascanu},
      year={2025},
      eprint={2410.01104},
      archivePrefix={arXiv},
      primaryClass={cs.LG},
      url={https://arxiv.org/abs/2410.01104}, 
}

@misc{zhang2025initializationcriticaltransformersfit,
      title={Initialization is Critical to Whether Transformers Fit Composite Functions by Reasoning or Memorizing}, 
      author={Zhongwang Zhang and Pengxiao Lin and Zhiwei Wang and Yaoyu Zhang and Zhi-Qin John Xu},
      year={2025},
      eprint={2405.05409},
      archivePrefix={arXiv},
      primaryClass={cs.LG},
      url={https://arxiv.org/abs/2405.05409}, 
}

@article{yip2024modular,
  title={Modular addition without black-boxes: Compressing explanations of {MLPs} that compute numerical integration},
  author={Yip, Chun Hei and Agrawal, Rajashree and Chan, Lawrence and Gross, Jason},
  journal={arXiv preprint arXiv:2412.03773},
  year={2024}
}

@article{zhao2025random,
  title={Random Scaling of Emergent Capabilities},
  author={Zhao, Rosie and Qin, Tian and Alvarez-Melis, David and Kakade, Sham and Saphra, Naomi},
  journal={arXiv preprint arXiv:2502.17356},
  year={2025}
}

@misc{akyurek2023learningalgorithmincontextlearning,
      title={What learning algorithm is in-context learning? Investigations with linear models}, 
      author={Ekin Aky{\"u}rek and Dale Schuurmans and Jacob Andreas and Tengyu Ma and Denny Zhou},
      year={2023},
      eprint={2211.15661},
      archivePrefix={arXiv},
      primaryClass={cs.LG},
      url={https://arxiv.org/abs/2211.15661}, 
}

@article{olah2020zoom,
  author = {Olah, Chris and Cammarata, Nick and Schubert, Ludwig and Goh, Gabriel and Petrov, Michael and Carter, Shan},
  title = {Zoom In: An Introduction to Circuits},
  journal = {Distill},
  year = {2020},
  note = {https://distill.pub/2020/circuits/zoom-in},
  doi = {10.23915/distill.00024.001}
}

@misc{li2025howlanguagemodelstrack,
      title={(How) Do Language Models Track State?}, 
      author={Belinda Z. Li and Zifan Carl Guo and Jacob Andreas},
      year={2025},
      eprint={2503.02854},
      archivePrefix={arXiv},
      primaryClass={cs.CL},
      url={https://arxiv.org/abs/2503.02854}, 
}

@article{kazemnejad2023impact,
  title={The impact of positional encoding on length generalization in transformers},
  author={Kazemnejad, Amirhossein and Padhi, Inkit and Natesan Ramamurthy, Karthikeyan and Das, Payel and Reddy, Siva},
  journal={Advances in Neural Information Processing Systems},
  volume={36},
  pages={24892--24928},
  year={2023}
}

@article{wurgaft2025context,
  title={In-Context Learning Strategies Emerge Rationally},
  author={Wurgaft, Daniel and Lubana, Ekdeep Singh and Park, Core Francisco and Tanaka, Hidenori and Reddy, Gautam and Goodman, Noah D},
  journal={arXiv preprint arXiv:2506.17859},
  year={2025}
}

@article{park2024competition,
  title={Competition dynamics shape algorithmic phases of in-context learning},
  author={Park, Core Francisco and Lubana, Ekdeep Singh and Pres, Itamar and Tanaka, Hidenori},
  journal={arXiv preprint arXiv:2412.01003},
  year={2024}
}

@article{zhong2023clock,
  title={The clock and the pizza: Two stories in mechanistic explanation of neural networks},
  author={Zhong, Ziqian and Liu, Ziming and Tegmark, Max and Andreas, Jacob},
  journal={Advances in neural information processing systems},
  volume={36},
  pages={27223--27250},
  year={2023}
}

@article{lindner2023tracr,
  title={Tracr: Compiled transformers as a laboratory for interpretability},
  author={Lindner, David and Kram{\'a}r, J{\'a}nos and Farquhar, Sebastian and Rahtz, Matthew and McGrath, Tom and Mikulik, Vladimir},
  journal={Advances in Neural Information Processing Systems},
  volume={36},
  pages={37876--37899},
  year={2023}
}

@article{zhou2023algorithms,
  title={What algorithms can transformers learn? a study in length generalization},
  author={Zhou, Hattie and Bradley, Arwen and Littwin, Etai and Razin, Noam and Saremi, Omid and Susskind, Josh and Bengio, Samy and Nakkiran, Preetum},
  journal={arXiv preprint arXiv:2310.16028},
  year={2023}
}

@article{awasthi2023improving,
  title={Improving length-generalization in transformers via task hinting},
  author={Awasthi, Pranjal and Gupta, Anupam},
  journal={arXiv preprint arXiv:2310.00726},
  year={2023}
}

@article{huang2026discovering,
  title={Discovering Interpretable Algorithms by Decompiling Transformers to RASP},
  author={Huang, Xinting and Bakalova, Aleksandra and Bhattamishra, Satwik and Merrill, William and Hahn, Michael},
  journal={arXiv preprint arXiv:2602.08857},
  year={2026}
}

@article{urdshals2025structure,
  title={Structure development in list-sorting transformers},
  author={Urdshals, Einar and Urdshals, Jasmina},
  journal={arXiv preprint arXiv:2501.18666},
  year={2025}
}

@article{moisescu2025geometry,
  title={On the geometry and topology of representations: the manifolds of modular addition},
  author={Moisescu-Pareja, Gabriela and McCracken, Gavin and Wiltzer, Harley and L{\'e}tourneau, Vincent and Daniels, Colin and Precup, Doina and Love, Jonathan},
  journal={arXiv preprint arXiv:2512.25060},
  year={2025}
}

@misc{nanda2023progressmeasuresgrokkingmechanistic,
  title         = {Progress measures for grokking via mechanistic interpretability},
  author        = {Nanda, Neel and Chan, Lawrence and Lieberum, Tom and Smith, Jess and Steinhardt, Jacob},
  year          = {2023},
  eprint        = {2301.05217},
  archivePrefix = {arXiv},
  primaryClass  = {cs.LG},
  url           = {https://arxiv.org/abs/2301.05217}
}

@inproceedings{weiss2021thinking,
  title     = {Thinking Like Transformers},
  author    = {Weiss, Gail and Goldberg, Yoav and Yahav, Eran},
  booktitle = {Proceedings of the 38th International Conference on Machine Learning},
  series    = {Proceedings of Machine Learning Research},
  volume    = {139},
  pages     = {11080--11090},
  year      = {2021},
  editor    = {Meila, Marina and Zhang, Tong},
  publisher = {PMLR},
  url       = {https://proceedings.mlr.press/v139/weiss21a.html}
}

@InProceedings{pmlr-v235-gatmiry24a,
  title     = {Simplicity Bias via Global Convergence of Sharpness Minimization},
  author    = {Gatmiry, Khashayar and Li, Zhiyuan and Reddi, Sashank J. and Jegelka, Stefanie},
  booktitle = {Proceedings of the 41st International Conference on Machine Learning},
  pages     = {15102--15129},
  year      = {2024},
  editor    = {Salakhutdinov, Ruslan and Kolter, Zico and Heller, Katherine and Weller, Adrian and Oliver, Nuria and Scarlett, Jonathan and Berkenkamp, Felix},
  volume    = {235},
  series    = {Proceedings of Machine Learning Research},
  month     = {21--27 Jul},
  publisher = {PMLR},
  url       = {https://proceedings.mlr.press/v235/gatmiry24a.html}
}

@InProceedings{pmlr-v235-tsoy24a,
  title     = {Simplicity Bias of Two-Layer Networks beyond Linearly Separable Data},
  author    = {Tsoy, Nikita and Konstantinov, Nikola},
  booktitle = {Proceedings of the 41st International Conference on Machine Learning},
  pages     = {48728--48767},
  year      = {2024},
  editor    = {Salakhutdinov, Ruslan and Kolter, Zico and Heller, Katherine and Weller, Adrian and Oliver, Nuria and Scarlett, Jonathan and Berkenkamp, Felix},
  volume    = {235},
  series    = {Proceedings of Machine Learning Research},
  month     = {21--27 Jul},
  publisher = {PMLR},
  url       = {https://proceedings.mlr.press/v235/tsoy24a.html}
}

@misc{du2024hierarchicalsimplicitybias,
  title         = {Hierarchical Simplicity Bias of Neural Networks},
  author        = {Du, Zhehang},
  year          = {2024},
  eprint        = {2311.02622},
  archivePrefix = {arXiv},
  primaryClass  = {cs.LG},
  url           = {https://arxiv.org/abs/2311.02622}
}

@inproceedings{shah2020pitfalls,
  title     = {The Pitfalls of Simplicity Bias in Neural Networks},
  author    = {Shah, Harshay and Tamuly, Kaustav and Raghunathan, Aditi and Jain, Prateek and Netrapalli, Praneeth},
  booktitle = {Advances in Neural Information Processing Systems},
  year      = {2020},
  eprint    = {2006.07710},
  archivePrefix = {arXiv},
  primaryClass  = {cs.LG},
  url       = {https://arxiv.org/abs/2006.07710}
}

\appendix
\section{The Impact of Initialization Scale}
\label{app:init_scale}

\subsection{The impact of changing initialization scale symmetrically}
\label{app:symm_init_scale}

\begin{figure}[t]
    \centering
    \includegraphics[width=\linewidth]{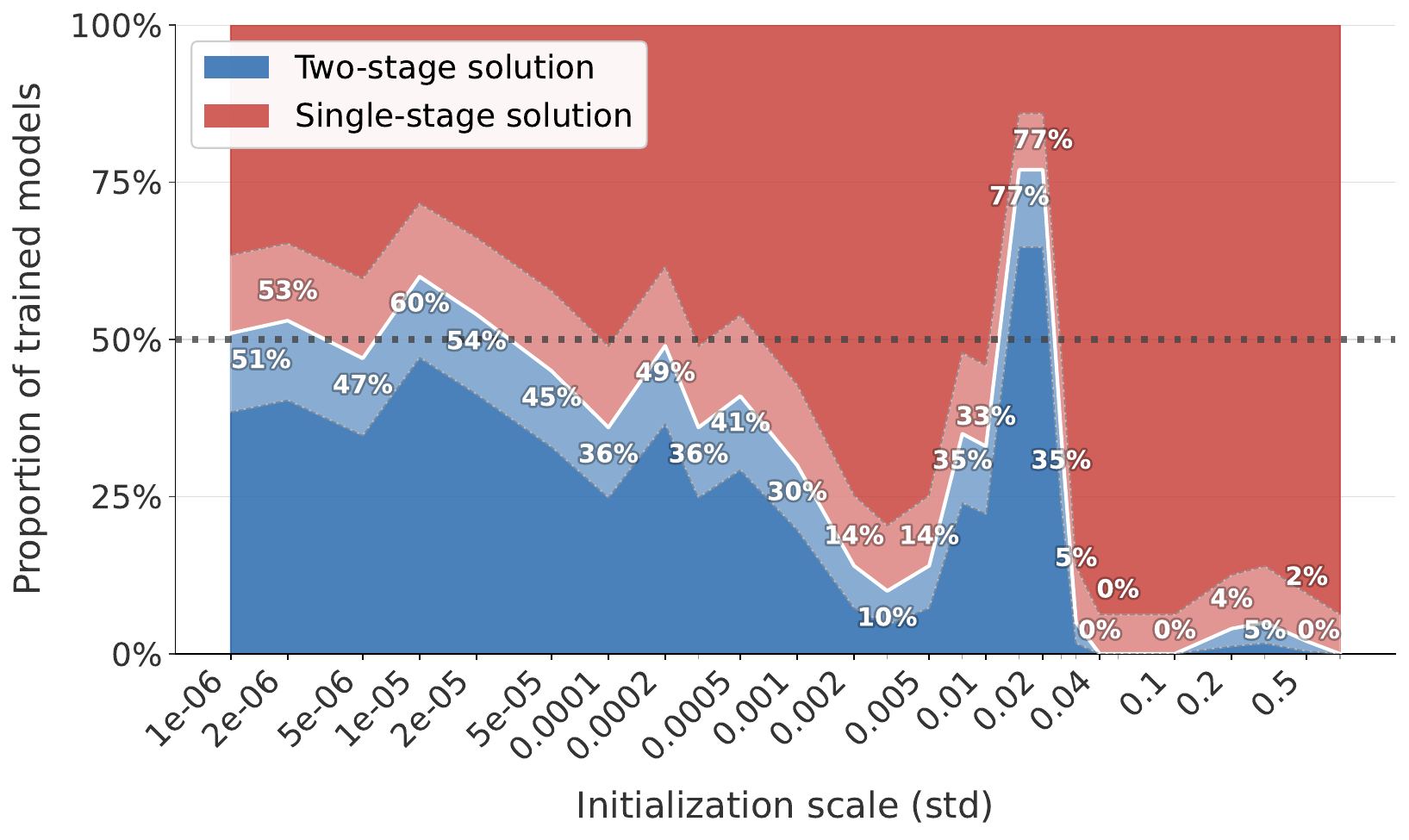}
    \caption{Fraction of seeds converging to the two-stage solution as a function of absolute initialization scale (100 seeds per condition, 99\% Wilson confidence intervals). The regime in which the two-stage solution dominates is narrow; outside it, nearly all seeds converge to single-stage solutions.}
    \label{fig:init_scale_figure}
\end{figure}

Section~\ref{sec:steering} showed that the relative initialization scale between layers steers which solution type emerges. Here we examine the effect of absolute initialization scale, where both layers are initialized at the same standard deviation~$\sigma$. Figure~\ref{fig:init_scale_figure} suggests that there may be three different regimes as $\sigma$ varies. For very small initialization scales, the outcome is approximately unbiased, where each solution type emerges roughly 50\% of the time, as if the initialization carries negligible information about which algorithm will be favored. For very large initialization scales, nearly all seeds converge to the single-stage (non-length-generalizing) solution. For intermediate initialization scales, we find that the behavior is less predictable, where some values of $\sigma$ favor two-stage solutions while others consistently favor single-stage solutions.

\subsection{Length Generalization of Asymmetrically Steered Models}
\label{app:steering_length_gen}

\begin{figure}[t]
  \centering
    \centering
    \includegraphics[width=\linewidth]{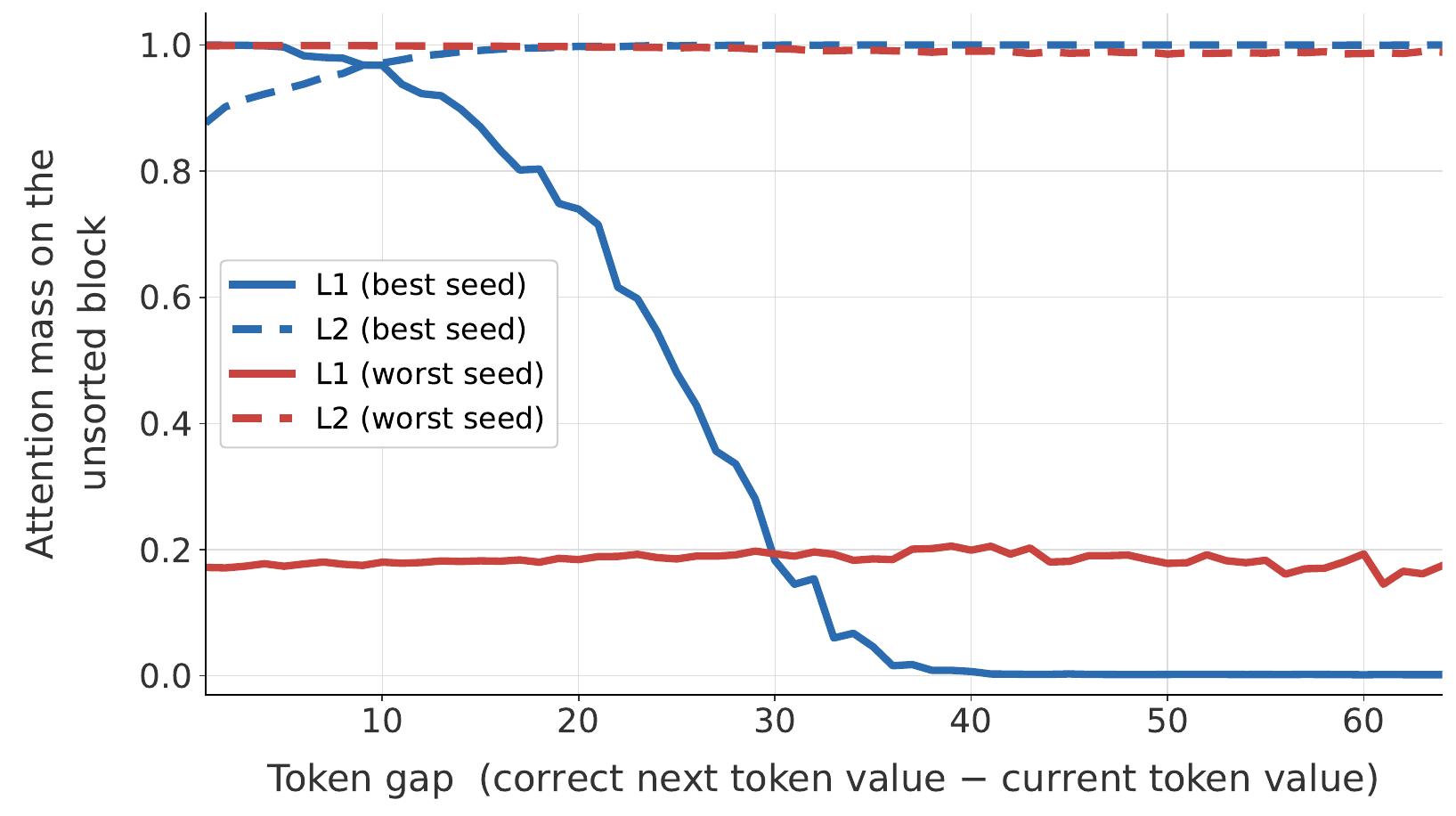}
  \caption{Best vs.\ worst length-generalizing steered seeds, where the second layer is initialized to be larger than the first ($\sigma_1 = 0.01, \sigma_2 = 0.013$). Of the models steered with asymmetric initialization and a larger layer~2 initialization scale than layer~1, we analyze the seed that length-generalized the best (blue) and the seed that length-generalized the worst (red). We find that the seed that length-generalized the best had similar specialization patterns across token gaps to two-stage models with symmetric initializations. However, there is indication that the model that generalized the worst, while being recorded as a two-stage model, relied more on the 2nd layer than the first.}
  \label{fig:oversteering}
\end{figure}

The bottom panel of Figure~\ref{fig:steering_figure} shows length generalization curves for all models in each condition, colored by solution type. At a qualitative level, two-stage models (blue) generalize, while single-stage models (red) do not, regardless of the difference in initial scale $\Delta$. However, we find a surprising sensitivity to relative initialization scale in that once the initialization of the second layer initialization is too strong, the model occasionally exhibits less cross-layer specialization (see Figure~\ref{fig:oversteering}), and has poorer length-generalization.

\section{The impact of vocabulary size and embedding dimension}
\label{app:vocab_embed}

\subsection{Varying vocabulary size}
\label{app:vocab_exp}
\begin{figure}[t]
    \centering
\includegraphics[width=\linewidth]{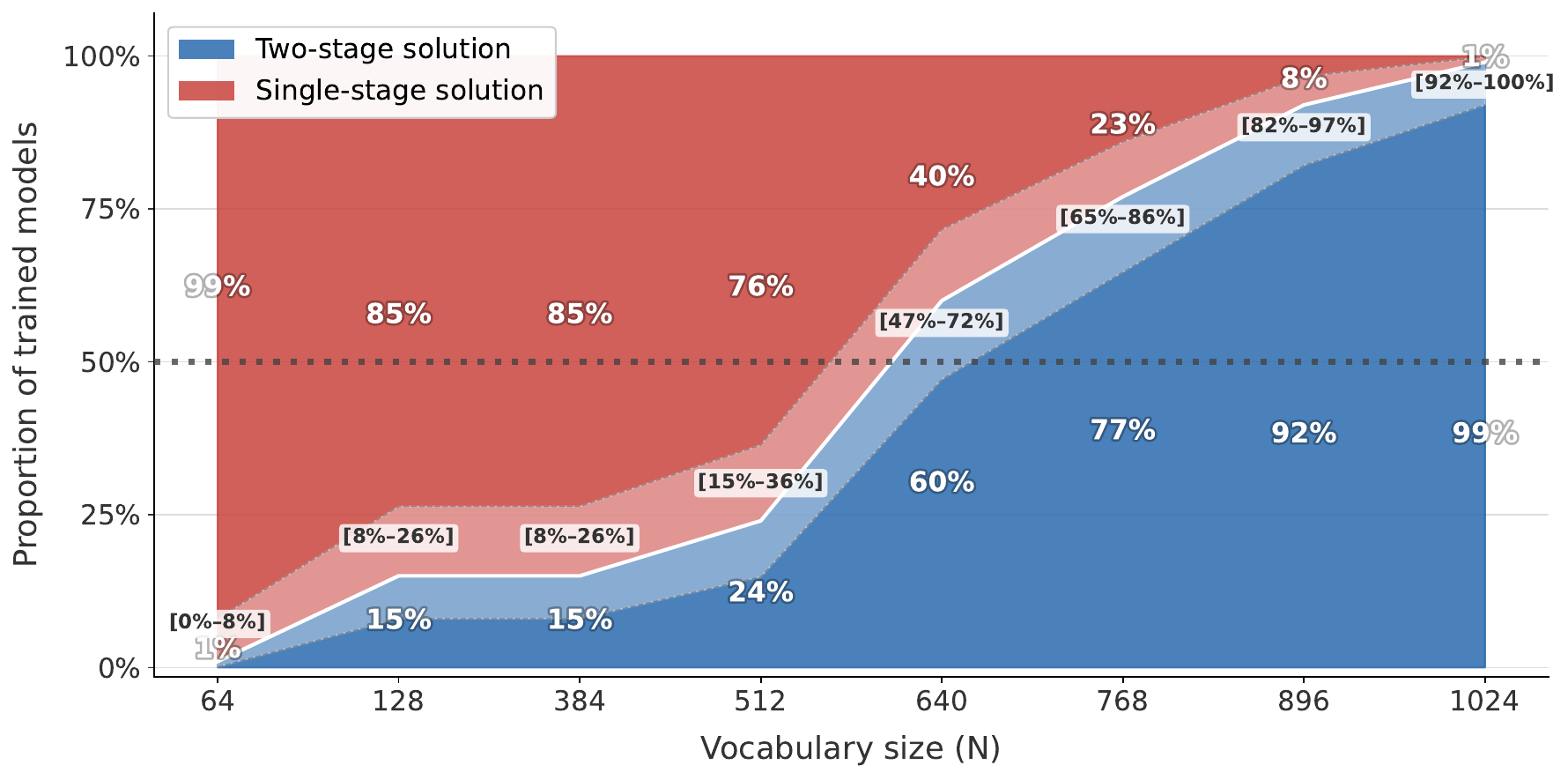}
  \caption{Solution type vs.\ vocabulary size~$N$. Proportion of seeds converging to each solution type as a function of vocabulary size~$N$ with sequence length $16$, reporting the probability of convergence to a two-stage model with 99\% confidence intervals, estimated with 100 seeds for each $N$. Larger vocabularies increasingly favor the two-stage solution.}
  \label{fig:varN}
\end{figure}

Section \ref{sec:task-difficulty} reports that larger vocabularies favor the two-stage solution while wider models favor the single-stage one. Figure \ref{fig:varN} varies the vocabulary size $N$ with all other hyperparameters fixed, using 100 seeds per vocabulary size. One potential interpretation of this phenomenon is that with larger vocabularies comes a harder discrimination problem, where with more possible token values, consecutive elements of the sorted output would be closer in embedding space, potentially making it more difficult for a single attention layer to reliably distinguish between them. Under this pressure, the network may be incentivized to use a more compositional solution from the start using the second layer.

\subsection{Varying embedding dimension}
\label{app:embed_exp}
\begin{figure}[t]
    \centering
    \includegraphics[width=\linewidth]{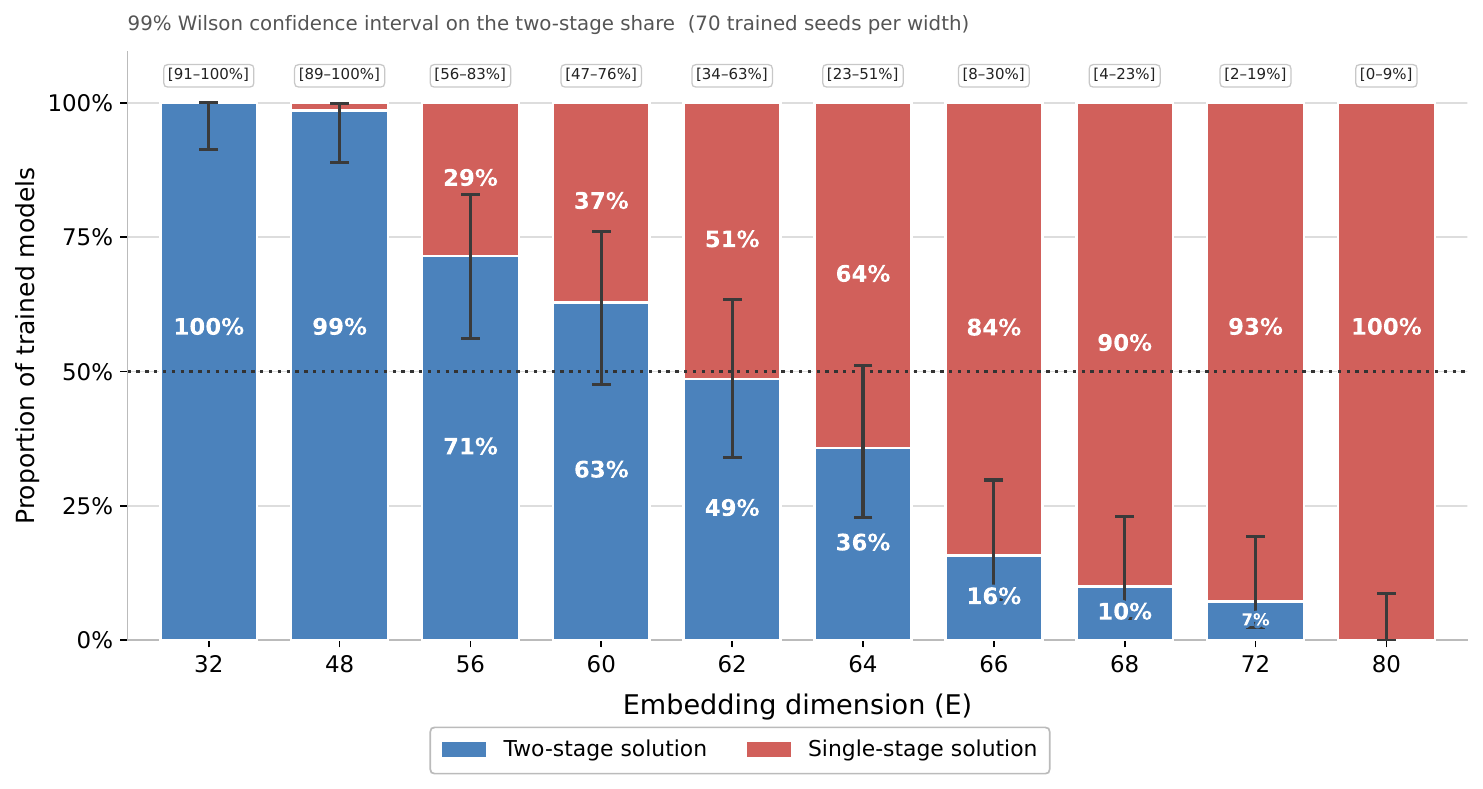}
  \caption{Solution type vs.\ embedding dimension~$E$. Proportion of seeds converging to each solution type as a function of embedding dimension~$E$, with sequence length $16$ and vocabulary size $N = 256$ held fixed, reporting the probability of convergence to a two-stage model with 99\% Wilson confidence intervals. Every bar is estimated from a common subsample of $70$ trained seeds---the largest number available at every width---so that all widths share a denominator and their intervals are directly comparable. Seeds that never learned the task are excluded, since ablating the second attention layer is only informative for models that solve the task. Wider models increasingly favor the single-stage solution, with the even-odds crossing between $E = 60$ and $E = 62$. This is the opposite direction to vocabulary size (Figure~\ref{fig:varN}) and consistent with the same explanation: the two-stage solution is recruited when the task is hard relative to the capacity available to solve it. Seeds are judged at their $60{,}000$-iteration checkpoint, matching the training length used elsewhere in the paper, although the cosine schedule was set for $100{,}000$ iterations so the learning rate has not fully annealed at that point. Repeating the classification at $100{,}000$ iterations changes no seed's assignment between the single-stage and two-stage classes; instead $16$ of $775$ seeds stop solving the task altogether, $14$ of them at $E = 32$, which moves each reported share by at most one point.}
  \label{fig:varE}
\end{figure}

One would expect capacity to act in the opposite direction as task difficulty. With a higher embedding dimension, the token representations have more mutually near-orthogonal directions available, so the query-key inner products that order the candidate tokens can be realized at a smaller embedding norm. Since the weights begin at a small initialization scale, a solution expressible at smaller norm should be easier for the optimizer to reach, and increasing the embedding dimension should therefore favor the single-stage solution. We vary the embedding dimension at a fixed vocabulary size (256) and sequence length ($16$), and estimate each width from a common sample of $70$ seeds that learned the task. Figure~\ref{fig:varE} confirms the prediction, showing that the two-stage share falls from every trained seed at embedding size $32$ to none at $80$.

\section{Training across varying input lengths}
\label{app:varying_k}

\begin{figure}[t]
  \centering
  \includegraphics[width=\linewidth]{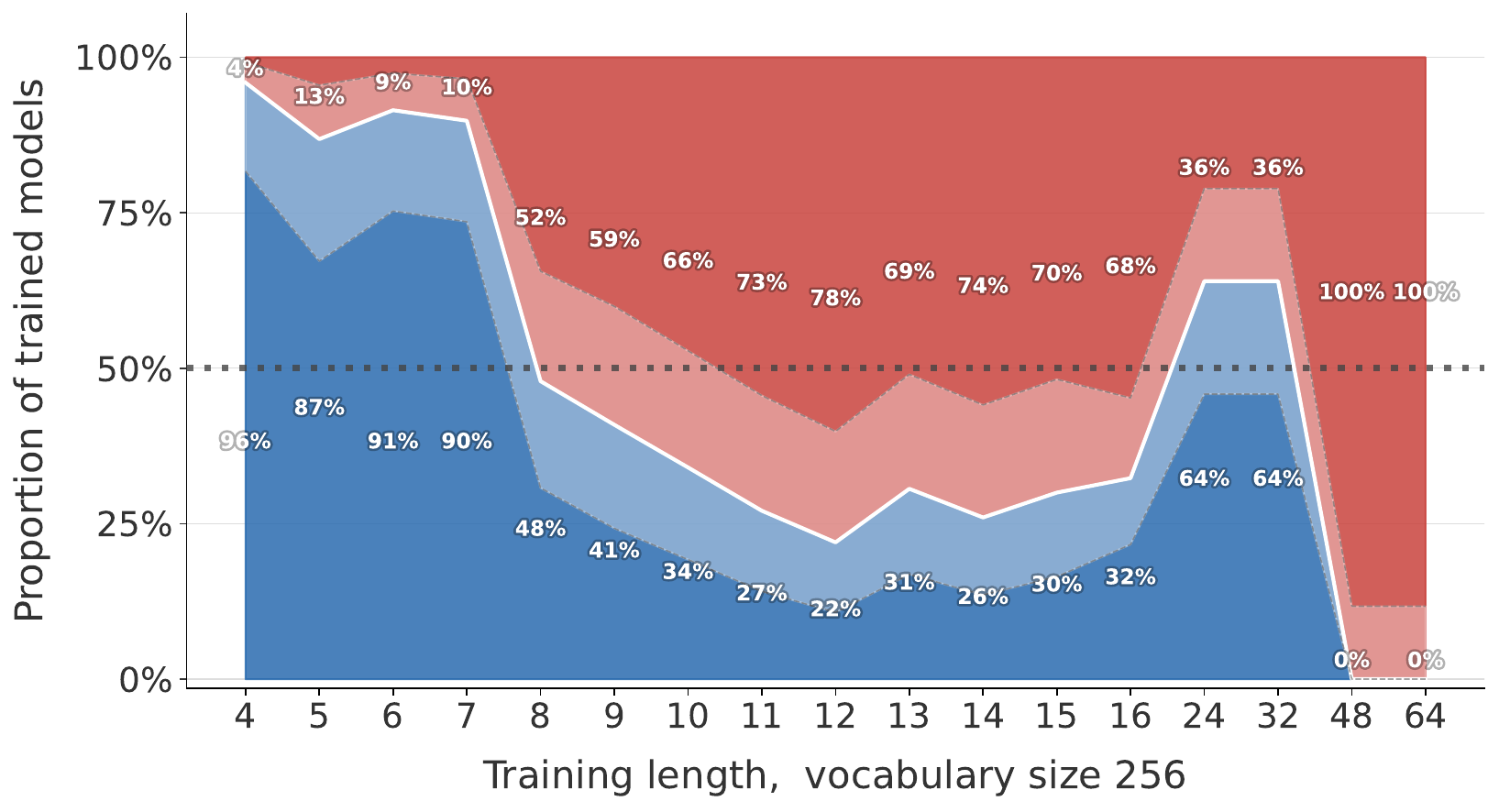}
  \caption{Solution type against training length. Proportion of trained models converging to the \textcolor{blue}{two-stage} and \textcolor{red}{single-stage} solutions at each training length $K$, with the vocabulary fixed at $N=256$, using $50$ seeds per choice of $K$. The lighter bands are $99\%$ Wilson confidence intervals on the probability of converging to a two-stage model.}
  \label{fig:varying_training_length}
\end{figure}

Our explanation of the two-stage model's length generalization rests on the gap distribution. When $K$ values are drawn from a vocabulary of size $N$, the mean gap between consecutive sorted values is $(N+1)/(K+1)$, so small gaps are rare at short training lengths and common at long ones. The training length therefore sets how often a model sees the inputs that the first attention layer of two-stage models specializes in (\S\ref{sec:specialization}). To test this, we vary the training length while holding the vocabulary fixed at $N=256$. We train $50$ seeds at each training length $K \in \{4,\dots, 16, 24, 32, 48, 64\}$ and classify each seed that learns the task based on reliance on the 2nd attention layer, as in the main text. We note that, interestingly, the share of two-stage models changes with training length non-monotonically (Figure \ref{fig:varying_training_length}). 

\begin{figure}[t]
  \centering
  \includegraphics[width=\linewidth]{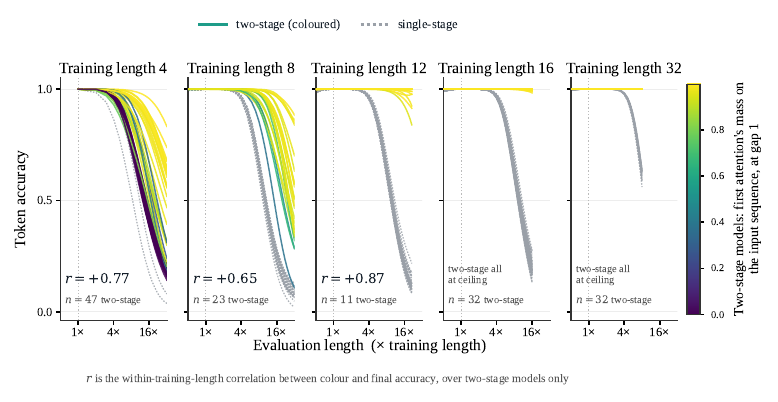}
  \caption{Length generalization of two-stage models, colored by first-layer attention on the input at gap $1$, as a proxy for layer-wise specialization, studied in Section \ref{sec:specialization}. Two-stage curves are colored by the attention mass that the first layer places on the input sequence for queries with gap size $1$ to their next value, measured at the training length. Single-stage curves are dotted gray. The value $r$ is the within-length Pearson correlation between color and token accuracy at length $256$.}
  \label{fig:length_gen_by_att1}
\end{figure}

\begin{figure}[t]
  \centering
  \includegraphics[width=\linewidth]{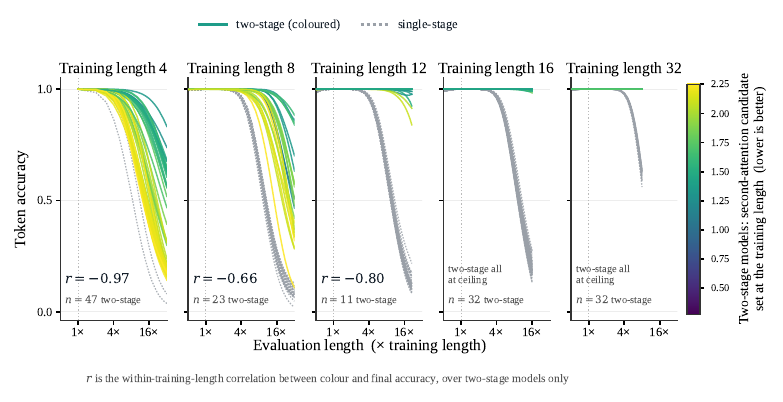}
  \caption{Length generalization of two-stage models, colored by the second attention's candidate set size. Color gives the average number of keys receiving second-attention weight above $0.04$ at training length. Single-stage curves are dotted gray. The value $r$ is the within-length Pearson correlation between color and token accuracy at length $256$.}
  \label{fig:length_gen_by_cand_set_size}
\end{figure}

\paragraph{Which two-stage models length-generalize?} At $K \geq 16$, every two-stage model generalizes essentially perfectly to the longest evaluation length. At shorter training lengths, however, generalization varies from one two-stage model to another and no seeds lead to perfectly length-generalizing models. We hypothesize this is because the frequency of tokens that have small gaps is much smaller when the training length is 4, so less expressivity is allocated to this. One property from the main text that predicts this variation is the amount of attention that the first layer places on the input sequence when the gap is $1$ (Figure \ref{fig:length_gen_by_att1}). Models whose first layer attends more to the input at small gaps (a property associated with exhibiting more gap-based specialization across layers) generalize further. In particular, the correlation with token accuracy at evaluation length $256$ is $r=+0.77$ at $K=4$, where we see the most variation in generalization abilities across seeds. We also find that models with smaller candidate sets generalize further (Figure \ref{fig:length_gen_by_cand_set_size}), achieving a similarly high anti-correlation between length-generalization abilities and candidate set size. Both results provide additional evidence that supports the two mechanisms of Section~\ref{sec:specialization} and Section~\ref{sec:attn2-qk-mechanism}.

\paragraph{Adversarial inputs across training lengths.}

\begin{figure}[t]
  \centering
  \includegraphics[width=\linewidth]{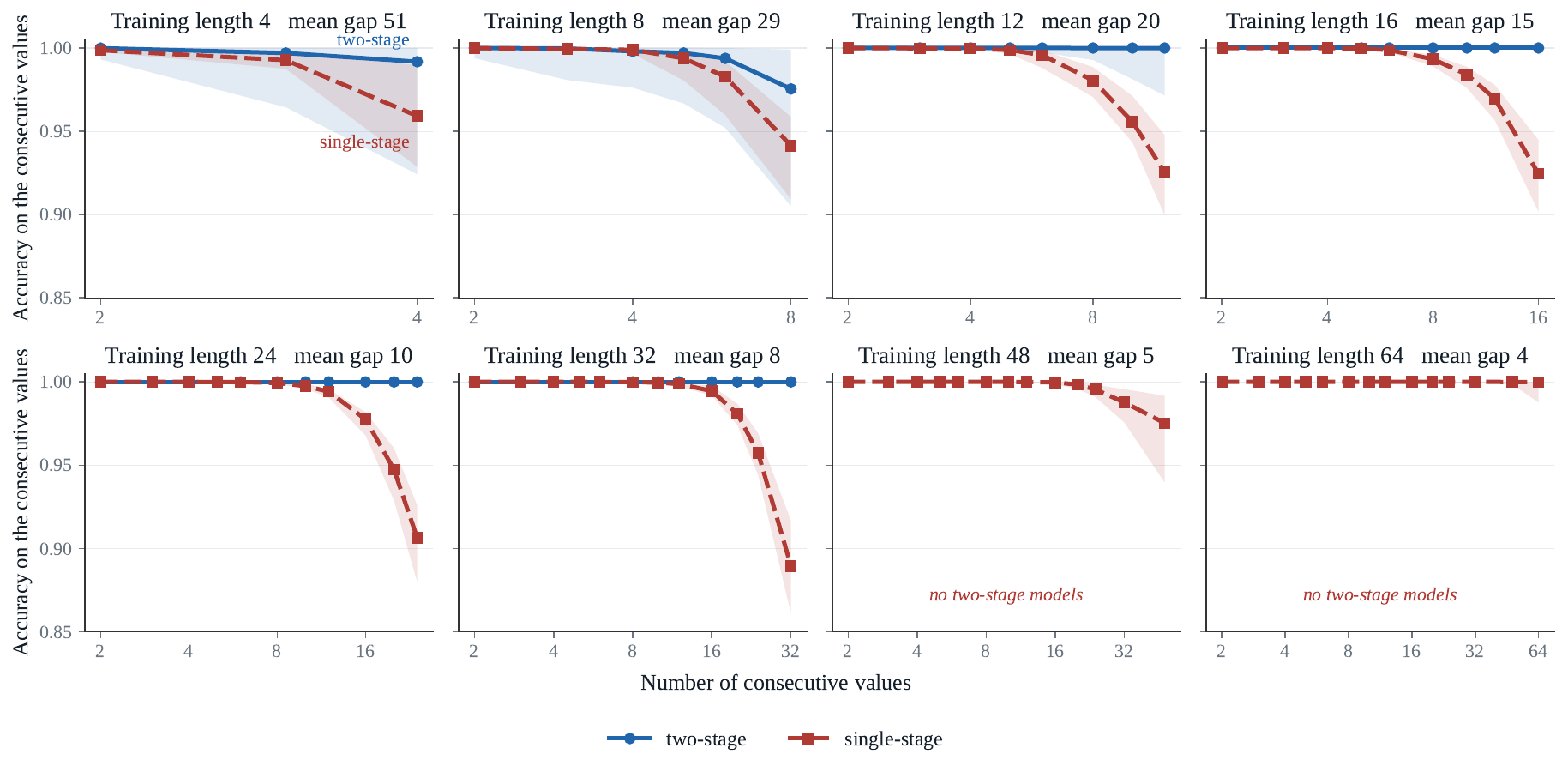}
  \caption{Small-gap adversarial examples across training lengths. Accuracy on runs of $n$ consecutive values}
  \label{fig:adv_ex_run_length}
\end{figure}

\begin{figure}[t]
  \centering
  \includegraphics[width=\linewidth]{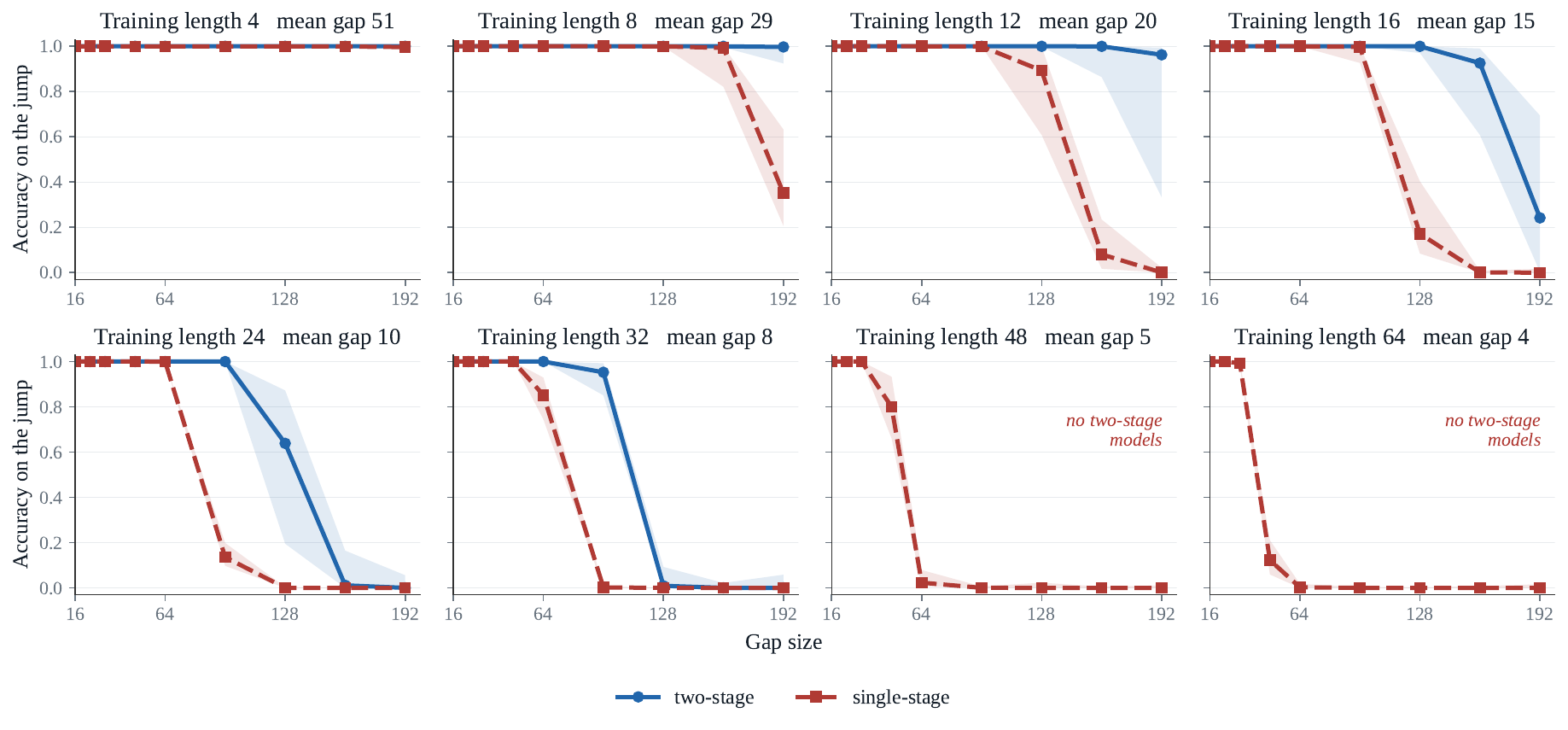}
  \caption{Large-gap adversarial examples across training lengths. We report accuracy on the prediction that crosses a gap of size $g$. Panel titles give the training length $K$ and the mean gap size seen during training for reference. Lines show the median over models and bands show the range. Gaps below 16 are handled perfectly and are omitted.}
  \label{fig:adv_ex_large_gap}
\end{figure}

As noted in the main text, we found that inputs consisting of permutations of consecutive sequences of numbers serve as adversarial examples to single-stage solutions, but are handled correctly by two-stage solutions. Here we inspect the extent to which models trained on different sequence lengths are exposed to similar kinds of adversarial examples. In particular, we plot the prevalence of small-gap adversarial examples and large-gap adversarial examples for single-stage and two-stage models across training lengths (Figures \ref{fig:adv_ex_run_length}, \ref{fig:adv_ex_large_gap}).

\emph{Small-gap adversarial examples} (Figure \ref{fig:adv_ex_run_length}).
We first consider inputs consisting of permutations of runs of $n$ consecutive integers, where the remaining $K-n$ values are sampled randomly outside the span of the run. We score accuracy only on the $n-1$ positions inside the run. For $K \geq 12$, two-stage models have an accuracy remaining at $1$ for every run length. This includes a fully consecutive sequence at $K=32$ (whose expected frequency in training is $10^{-39}$), where single-stage models score $0.89$. Single-stage models become more robust to smaller gaps as $K$ grows and small gaps become more common in training, and at $K=64$ even single-stage models score $1.00$ on a fully consecutive sequence. However, for very large training lengths, while you see small gaps more often, large gaps become increasingly uncommon.

\emph{Large-gap adversarial examples} (Figure \ref{fig:adv_ex_large_gap}).
Each large-gap adversarial example contains a single adjacent pair $v, v+g$ with the interval between them left empty, and we score accuracy on the one prediction that must jump from $v$ to $v+g$. Every model handles gaps up to some size and then fails abruptly. The maximum gap that can be predicted accurately shrinks as the training length grows, following the mean gap seen in training, and two-stage models typically handle larger gaps than single-stage models trained at the same length. We believe that this is further attributed to gap-wise specialization across layers. Together, these two families of adversarial examples provide additional evidence suggesting that the gap distribution that the model is trained on impact what inputs a model can handle. Longer training lengths make models robust to small gaps but vulnerable to large ones, and shorter training lengths make models more accurate at large gaps but worse with small ones, leading to the worse length-generalization we see for models trained at small lengths.

\section{Sorting in Deeper Transformers and Multi-headed Transformers}
\label{app:4-layer-section}

Given the existence of single-stage solutions to sorting, we know that a single attention layer with a single head is sufficient to sort. As a result, our main experiments training two-layer models are partially motivated by training a model with one more layer than it needs, where we find that whether it uses the second layer depends on the seed. Here, we ask whether the same holds when the model has more spare capacity, either as additional layers beyond 2 (Appendix \ref{app:4-layer}) or as additional heads within one layer (Appendix \ref{app:multihead}).

\subsection{Four-layer transformers}
\label{app:4-layer}

We also train deeper transformers on the sorting task. While two-layer transformers learned one of two solution types (the single-stage or two-stage solution), we find that four-layer transformers learn one of four solution types, where in each type, different attention layers are removable. 

\paragraph{Setup.} We use the same architecture as in Section \ref{sec:setup} (one head per layer, embedding dimension $64$, and no positional embeddings), with four transformer layers instead of two. They are trained to sort sequences of length 16 from a vocabulary of size 128, using AdamW with a learning rate of $0.02$ and a batch size of $4096$, and a cosine schedule with $200$ warm-up steps for $20000$ iterations. We compare two initialization conditions on the same $50$ seeds, one with the same initialization scale for every layer ($0.01$) and one with a lower scale for every layer after the first ($\sigma_1 = 0.01$ and $\sigma_2 = \sigma_3=\sigma_4 = 0.005$).

\paragraph{Solution types.}
To find which layers a model needs, we zero attention layers of trained models and re-evaluate. A series of layers counts as removable if the full sequence accuracy stays at or above $0.999$. However, the number of removable layers is very robust to this threshold number. We define a model's solution type by the number of layers it cannot do without, including one (only the first layer is needed), two (only the first two), three (only the first three), and four (every layer is needed). We find that the necessary layers always form such a contiguous sequence of necessary layers (i.e. only the fourth layer can be removed if one is only removing one attention layer).

\begin{figure}[t]
  \centering
\includegraphics[width=\linewidth]{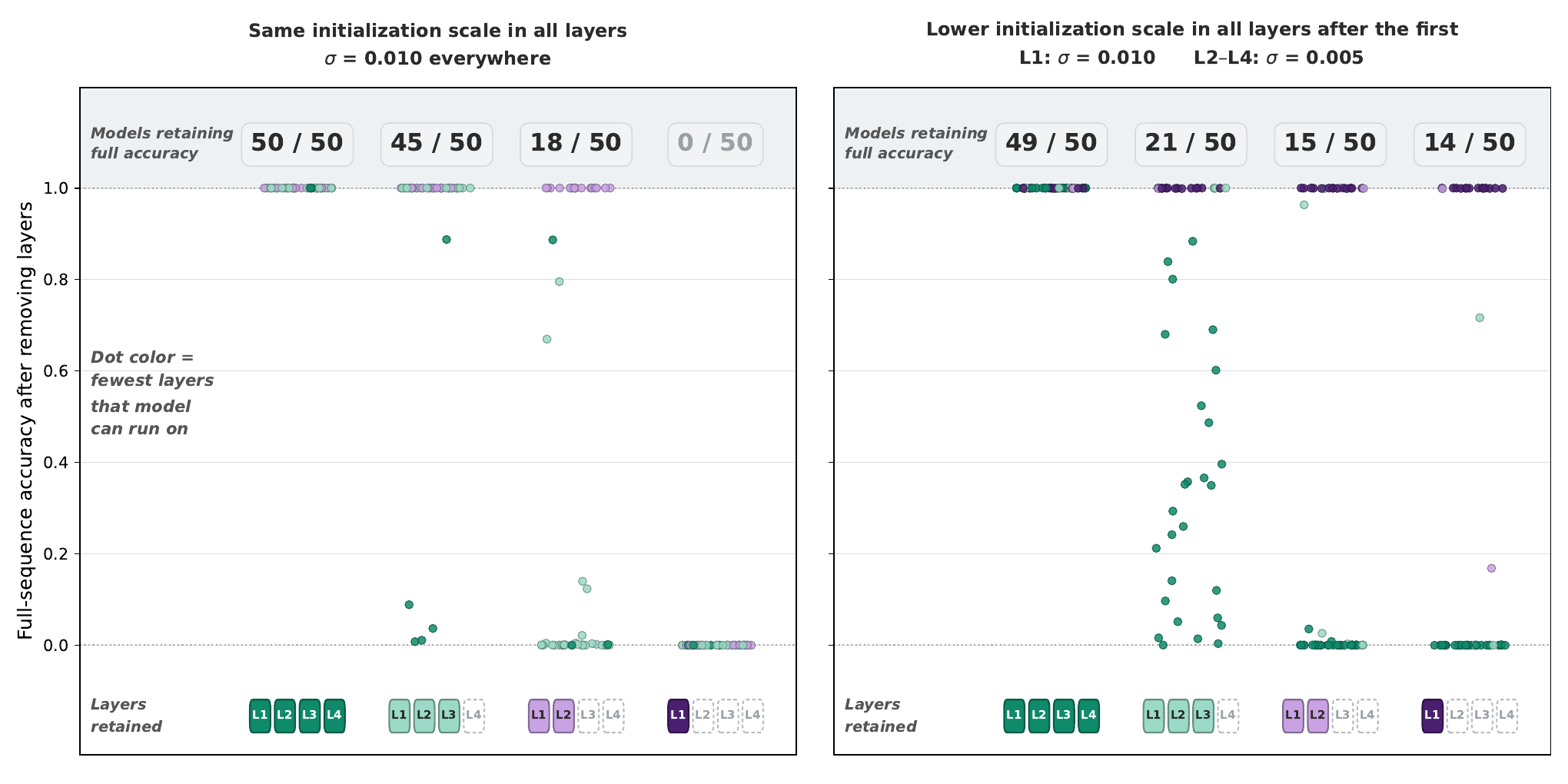}
  \caption{The multiplicity of solution classes, and their steerability, persists at depth four. Each column removes a deeper suffix of attention layers simultaneously (diagram at bottom) and re-evaluates. Plotted points are individual models, colored by the fewest leading layers the model can run on. Under equal initialization scales, no model reduces to a single attention layer, even though one would suffice. Lowering the scale of layers $2$ to $4$ steers $14$ of the $50$ models to a one-layer solution.}
  \label{fig:4-layers}
\end{figure}

\paragraph{Initialization steers how many layers are used.}

Figure \ref{fig:4-layers} displays the distribution of solution types for 4-layer transformers trained on the sorting task. Under equal scales, some seeds require four, three, and two layers to sort, and none of the seeds reduce to just a single layer, even though one would be sufficient. Lowering the scale of layers 2 through 4, however, produces 14 models that run on the first layer alone. This mirrors the two-layer setting (Section \ref{sec:steering}), where making the first layer relatively larger at initialization favors the single-stage solution.

\begin{figure}[t]
  \centering
  \includegraphics[width=\linewidth]{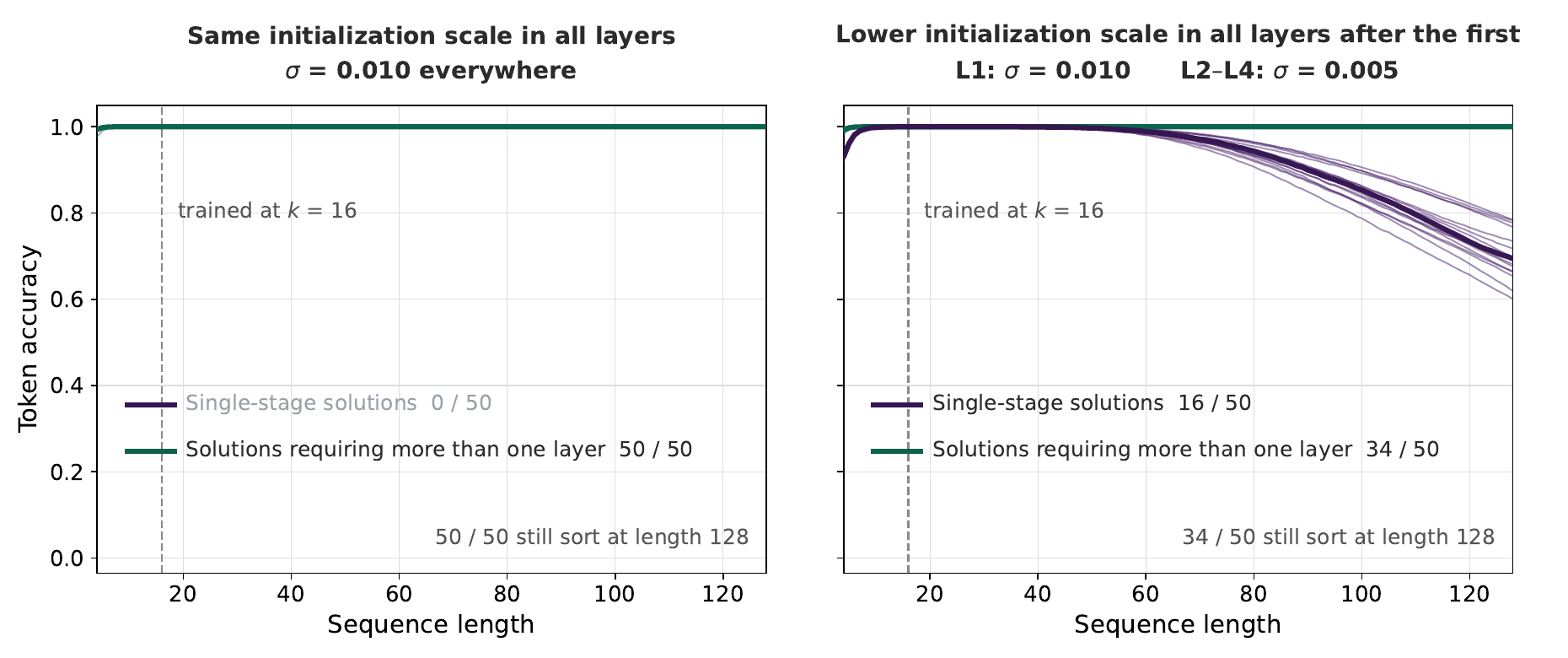}
  \caption{Mirroring our results for transformers of depth 2, when transformers of depth 4 are trained to sort sequences of numbers, only the solutions that utilize at least 2 layers are able to length-generalize.}
  \label{fig:4-layers-length-gen}
\end{figure}

\paragraph{Length generalization still requires at least two layers.}

% to-do, I want to make sure we don't forget any figure references in the text

Like with the two-layer set-up, in the four-layer set-up we call a model single-stage when it works when all attention layers but the first are ablated. Mimicking the two-layer case, single-stage models also fail to length-generalize (Figure \ref{fig:4-layers-length-gen}), while any seed using more than one layer succeeds in length generalization (like the two-stage solutions in the two-layer set-up).

We do not attempt to characterize three- and four-layer solutions mechanistically. A cursory exploration led us to find that the mechanisms identified for two layers do not generalize neatly to more layers. In particular, there does not appear to be a clean layer-wise specialization of tokens, and there does not appear to be a more general divide and conquer strategy that appears as solutions get deeper (i.e. candidate sets do not get iteratively smaller).

\subsection{Multiple heads per layer}
\label{app:multihead}

We next give the model spare capacity within a layer instead of adding additional layers. While with two layers, whether the second layer is used depends on the seed, we find that when we extend to multiple heads, all seeds require all heads for performance and the variation across seeds exists only for whether or not the second layer is used rather than whether one or multiple heads are used.

\paragraph{Setup.} We train models with vocabulary size 256 and a training length of 16 using the same training recipe of Section \ref{sec:setup}, varying both the number of layers and the number of heads per layer $H$. The $H$ heads of a layer split into heads of dimension $d_{\text{model}}/H$. We remove a head by zeroing its output before the layer's output projection. Heads within a layer are interchangeable, so for each model we remove the head whose removal results in the lowest loss in accuracy.

% multiple head figures 
\begin{figure}[t]
  \centering
  \includegraphics[width=\linewidth]{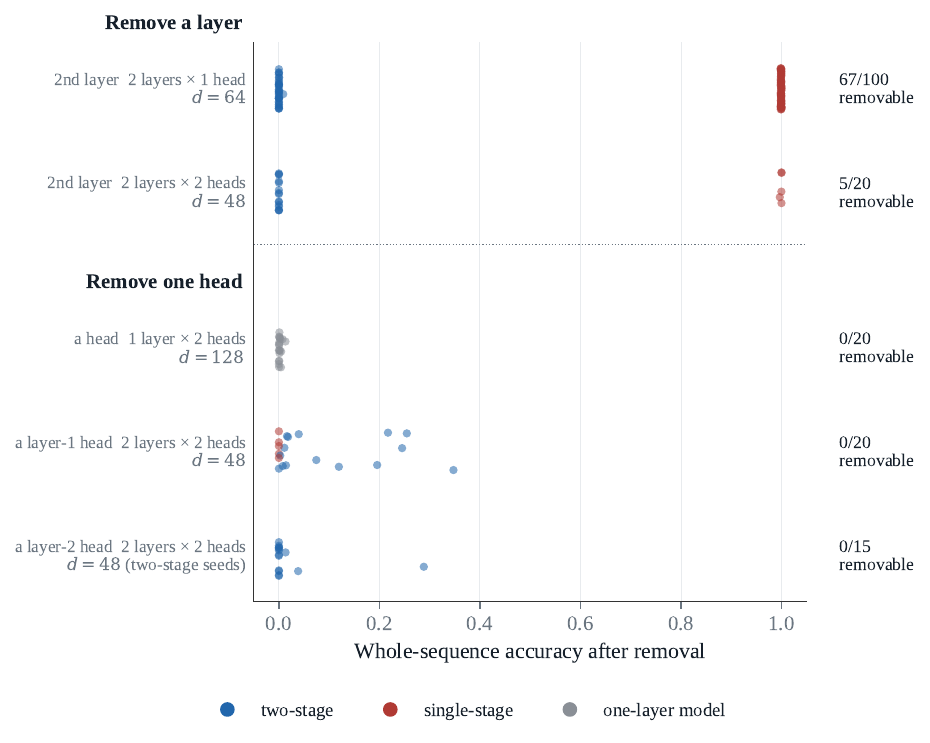}
  \caption{Whether a second layer is used varies by seed, but when we train models with multiple heads, all heads are always used when the layer is being used. We report sequence accuracy after removing one unit from each model in each row. For heads, we remove the head whose removal costs the model the least. On the right side of the plot, we display the number of models that keep at least $95\%$ of their accuracy. \textbf{Top.} Removing the second attention layer leaves some seeds unaffected and breaks others, both with one head per layer and with two. \textbf{Bottom.} Removing a single head breaks every model, whether it is one of two heads in a one-layer model (where each head alone would be enough to perfectly solve the task) or a head of either layer in the two-head, two-layer models.}
  \label{fig:multi_heads_removability}
\end{figure}

\paragraph{Whether a second layer is used varies by seed, while a second head is always used across seeds.} Figure \ref{fig:multi_heads_removability} compares the two kinds of spare capacity (spare layers and spare heads). The second layer is removable without loss for some seeds in both single-head models and two-head models, depending on the seed. In one-layer models with two heads, with $d_{\text{model}} = 128$, where either head would be large enough on its own, ablating a head causes all $20$ models to lose most of their accuracy. The same holds for heads of both layers in the two-head, two-layer models at $d_{\text{model}}=48$ (excluding the $5$ single-stage seeds, where one would be able to ablate any of the second layer's heads since the entire attention layer is removable). Thus, we find that when an attention layer is used by the model, all of its heads are used. We note that this holds for up to three heads per layer. With $4$ or $8$ heads per layer, we note that removable heads in a used second layer become more common.

\begin{figure}[t]
  \centering
  \includegraphics[width=\linewidth]{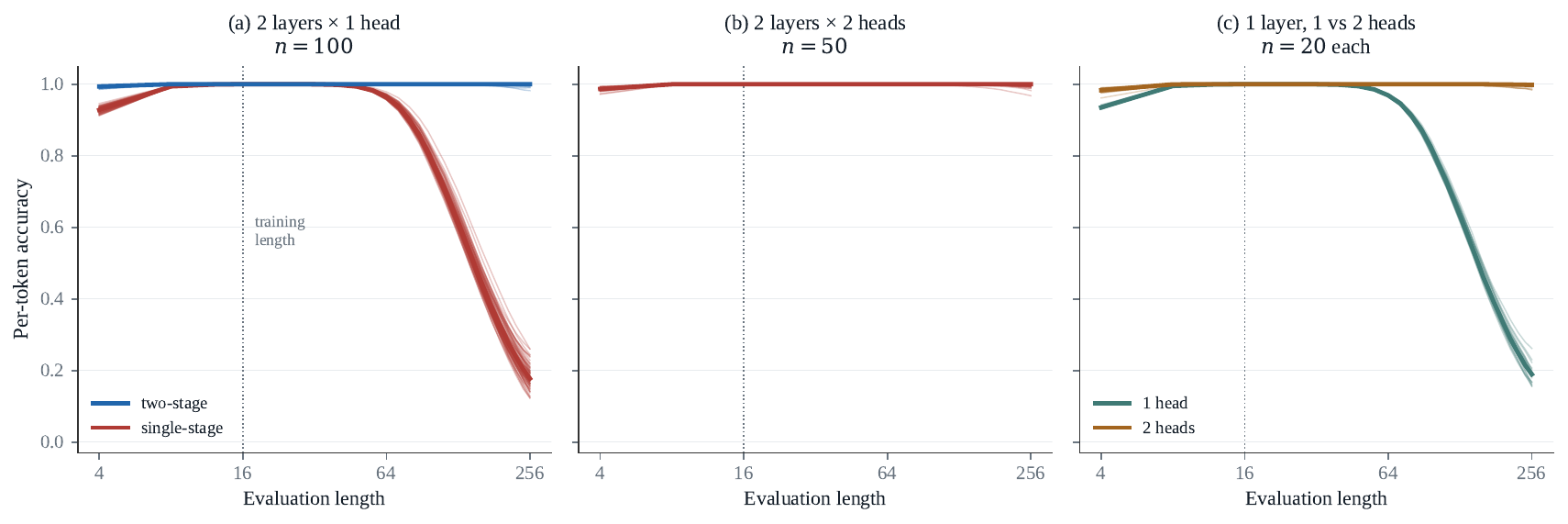}
  \caption{A second head prevents the length-generalization failure. We evaluate token accuracy across input lengths for models with embedding dimension $64$ that are trained on length-16 inputs. \textbf{(a)} Two-layer models with one head per layer. Single-stage models (red) fail at long lengths, while two-stage models (blue) do not. \textbf{(b)} Two-layer models with two heads per layer. This set-up is chosen so that every model is single-stage; however, we find that all single-stage models length-generalize when trained with 2 heads per layer. \textbf{(c)} We train both one-layer models with one head and one-layer models with two heads and find that only those with two heads length-generalize.}
  \label{fig:multi_heads_length_gen}
\end{figure}

\paragraph{A second head prevents the length-generalization failure.}
We find that the length-generalization failure of single-stage solutions in the main text is specific to one head per layer (Figure \ref{fig:multi_heads_length_gen}). With one head, single-stage models fail at long lengths while two-stage models do not (Figure \ref{fig:multi_heads_length_gen} (a)). With two heads, every model is single-stage, yet all of them length-generalize (Figure \ref{fig:multi_heads_length_gen} (b)). The same holds for one-layer models, which have no second stage at all (Figure \ref{fig:multi_heads_length_gen} (c)). We find that this is not particular to one width, as models with two heads and two layers length-generalize at every width and initialization scale that we trained, regardless of their solution type.

\section{Second Attention QK Scores Depend Predominantly on MLP1 Output}
\label{app:attn2-mlp1-dependence}

\begin{figure*}[t]
  \centering
  \includegraphics[width=\textwidth]{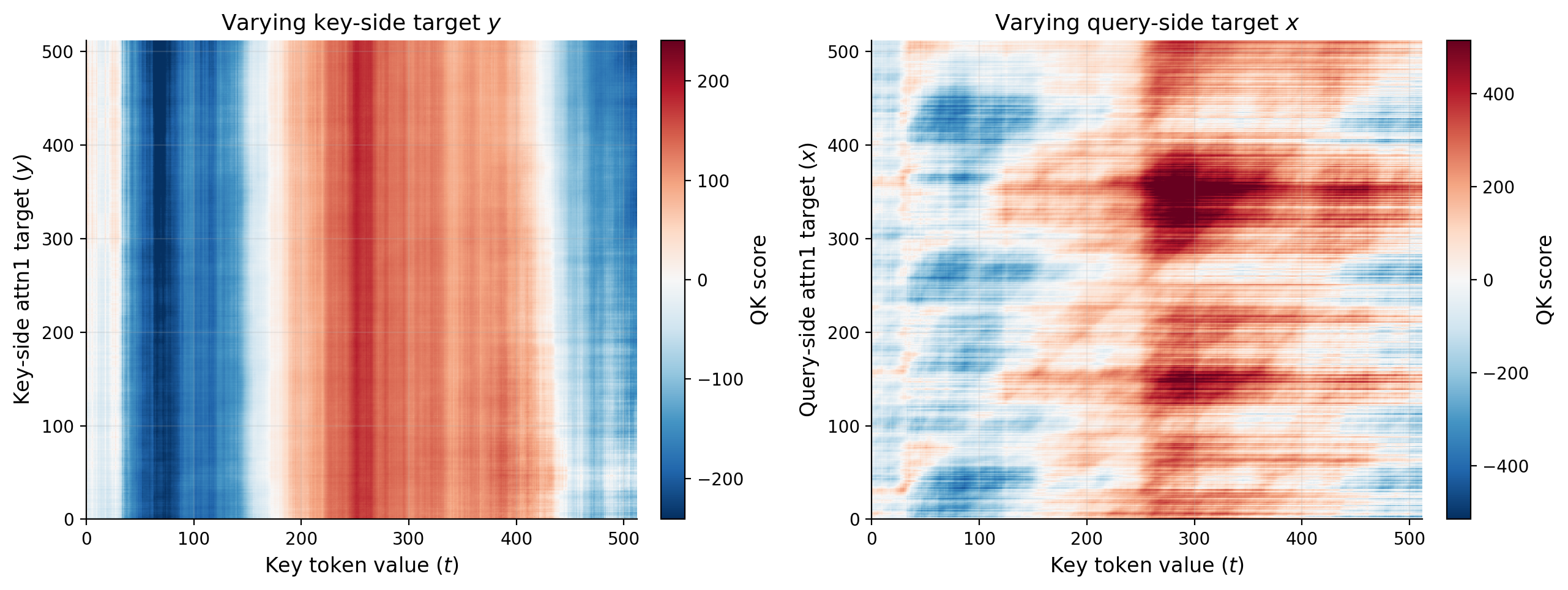}
  \caption{Second attention QK score heatmaps for query token $z = 250$.
    \textbf{Left:}~The score is nearly invariant to the key-side first attention
    target~$y$, showing uniform horizontal bands.
    \textbf{Right:}~The score is strongly modulated by the query-side
    target~$x$: as $x$ increases, a high-score band emerges in $[z, x]$.}
  \label{fig:qk-heatmap-asymmetry}
\end{figure*}

In the main text (\S\ref{sec:attn2-qk-mechanism}) we write the second attention score as a function $s(z, x, t, y)$ of only four quantities: the query token, the key token, and the first attention's target on each side. This simplification is justified because, in the majority of two-stage checkpoints, the second-layer
attention scores are predominantly determined by the first-layer MLP output
alone. In particular, replacing the second attention's input with the first MLP's output causes negligible changes in both the attention distribution and full-model accuracy. We verify this across a grid of sequence lengths $K \in \{16, 32\}$ and
vocabulary sizes $N \in \{128, 256, 512, 1024\}$ with 5~random seeds each (40~checkpoints
total), of which 15 are classified as two-stage models (i.e., full-sequence
accuracy drops to near zero when the second attention is ablated). We exclude two outlier
checkpoints at vocabulary size $1024$ discussed below, leaving 13 for detailed analysis.
For each checkpoint we evaluate on 5{,}000 randomly sampled sequences,
measuring over all output positions. We perform two complementary measurements.

\begin{figure}[h]
  \centering
  \includegraphics[width=\linewidth]{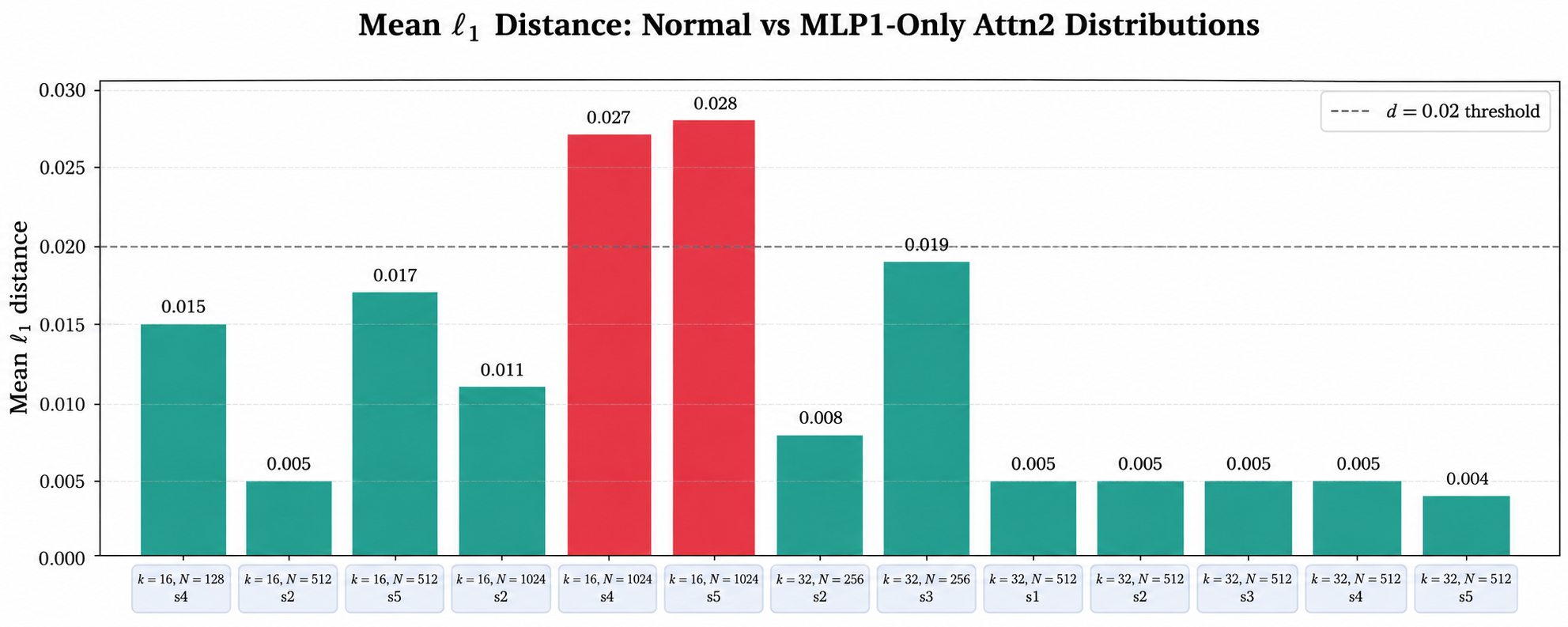}
  \caption{Mean $\ell_1$ distance between normal and MLP1-only second attention
    distributions for 13 two-stage checkpoints (outliers excluded).}
  \label{fig:probl1distance}
\end{figure}

\paragraph{Distributional similarity.}
We compare the second attention's probability distribution computed from the full residual
stream ($\mathbf{e} + \mathbf{a}_1 + \mathbf{m}_1$) against the distribution
computed from only the MLP1 output ($\mathbf{m}_1$), with embeddings and the first attention's
output zeroed out. We report the per-position $\ell_1$ distance
$d = \sum_j |p_{\text{normal}}(j) - p_{\text{mlp1-only}}(j)|$, which ranges
from~$0$ (identical) to~$2$ (disjoint). All 13 checkpoints exhibit mean
$\ell_1$ distances below $0.03$, with all five seeds at sequence length $32$ and vocabulary size $512$
yielding $d \approx 0.005$ (Figure~\ref{fig:probl1distance}).

\begin{figure}[h]
  \centering
  \includegraphics[width=\linewidth]{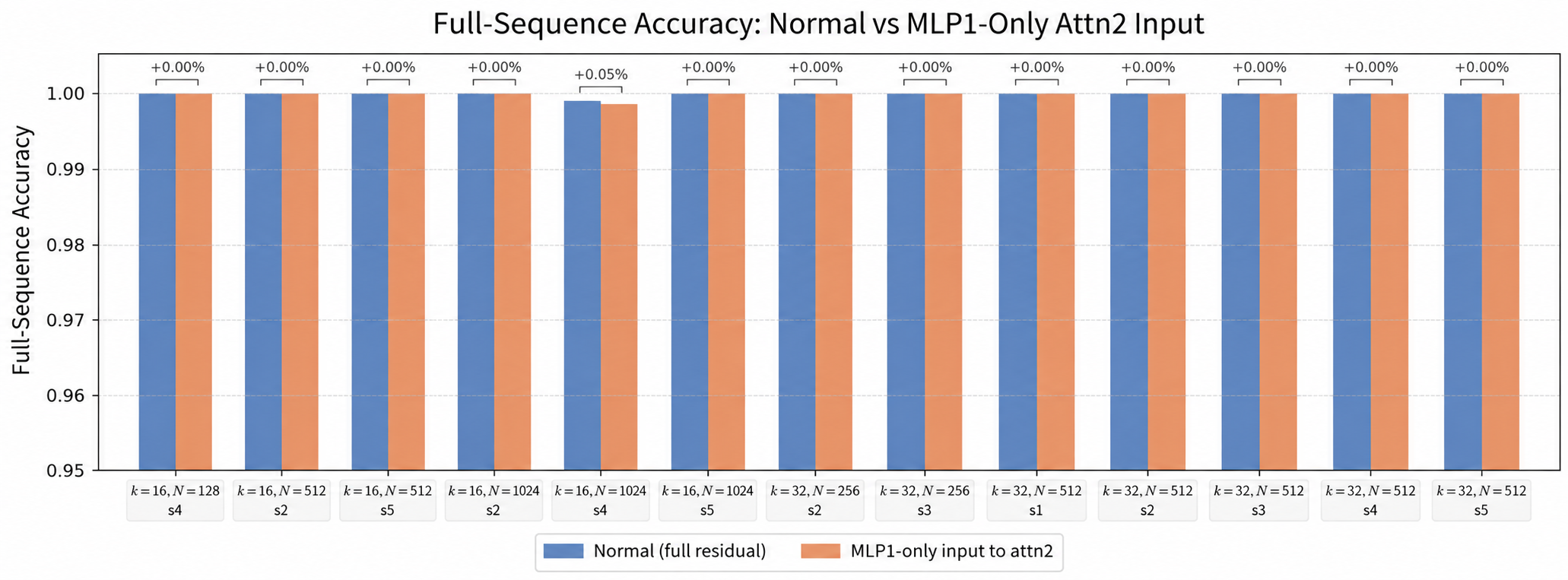}
  \caption{Full-sequence sorting accuracy under normal versus MLP1-only second attention
    input. The accuracy drop is zero for 12 of 13 checkpoints and $0.05\%$ for
    the remaining one. 
    % \nathan{This doesn't need to be a figure in my opinion as we could just write this result in text directly.}
    }
  \label{fig:attn2-acc-mlp1only}
\end{figure}

\paragraph{Full-model accuracy.}
As a functional test, we also measure the model's full-sequence sorting
accuracy when the second attention's input is restricted to MLP1 output only (while all other
components, including MLP2 and the final readout, receive the normal residual
stream augmented with this modified second attention output).
Figure~\ref{fig:attn2-acc-mlp1only} shows that the accuracy drop is zero or
negligible across all 13 checkpoints (maximum drop $0.05\%$), confirming that
the embedding and first attention components of the residual stream carry no information
that the second attention functionally relies on.

\paragraph{Outliers.}
Two of the 15 two-stage checkpoints at vocabulary size $1024$
(sequence length $16$, seed~3, $d = 0.133$; sequence length $32$, seed~5, $d = 0.299$)
deviate from this pattern, showing that the second attention can also learn to exploit other
residual stream components. Since other seeds at vocabulary size $1024$ do not exhibit this
behavior, it appears to be seed-dependent rather than determined by vocabulary
size. We note that the MLP1-only dependence is a common empirical feature of
two-stage models in our setting, not a necessary architectural property.

\section{Smoothness Advantage of the Second Attention over the First}
\label{app:smoothness}

\begin{figure}[h]
  \centering
  \includegraphics[width=\linewidth]{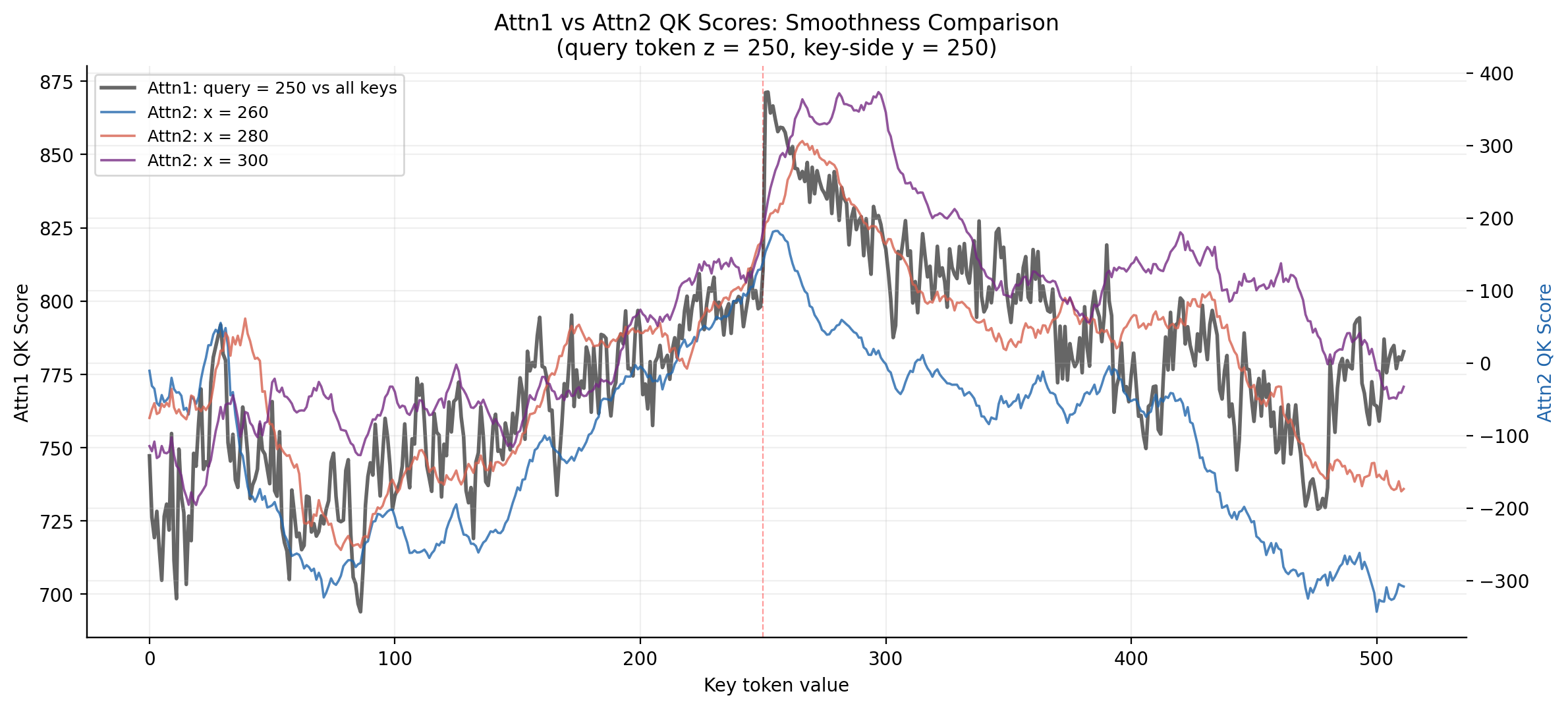}
  \caption{First attention QK score (black, left axis) versus second attention QK score slices
    (colored, right axis) as a function of key token value. The second attention's scores are
    smoother within their respective windows, reflecting the benefit of the
    narrower search range.}
  \label{fig:l1-l2-smoothness}
\end{figure}

Figure~\ref{fig:l1-l2-smoothness} overlays the first attention's QK score (for query token
$250$ vs.\ all keys) with the second attention's QK scores at three query-side targets.
The first attention's curve is noisy across the full vocabulary range, reflecting the
difficulty of maintaining monotonicity over $512$ tokens. The second attention's
curves, by contrast, are visibly smoother in their respective windows---the
narrower the window, the easier it is for the learned embeddings to maintain a
clean monotonic profile.

\section{How second attention uses the hint from the first attention when its score is nontrivial on multiple tokens} %Windowed Monotonicity Under Split First Attention}
\label{app:split-attn1}

The windowed monotonicity analysis in \S\ref{sec:attn2-qk-mechanism} assumes
the first attention places its full weight on a single token~$x$. In practice, the first attention
distributes weight over a small candidate set (typically 2--3 tokens above an
attention threshold of $0.04$). To test the robustness of our hypothesis in section~\ref{sec:attn2-qk-mechanism} on how second attention uses the contextual information from the first attention, instead of assuming the first attention is only concentrated on one token, we generalize it by studying the case when the first attention is concentrated on a set of $n$ equidistant numbers $\{x_{min}, x_{min} + \delta, \dots, x_{min} + (n-1)\delta\}$ %a uniform split among $n$ tokens
spaced by~$\delta$. 
Consistent with our notation in section~\ref{sec:attn2-qk-mechanism}, the query embedding of the second-layer attention is now given by 
\begin{align*}
    &\mathbf{q}(z,x,n,\delta) = W_Q^{(2)} \operatorname{LN}\!\bigl(
\operatorname{MLP}_1\!\bigl(\operatorname{LN}(e_z + V_{split}(x,n,\delta))\bigr)\bigr),
    %&\mathbf{q}(z,x) = W_Q^{(2)} \operatorname{LN}\!\bigl(
%\operatorname{MLP}_1\!\bigl(\operatorname{LN}(e_z + V_{split}(x,n,\delta))\bigr)\bigr),
\end{align*}
where
\[
  V_{\text{split}}(x, n, \delta)
  = \frac{1}{n}\sum_{i=0}^{n-1} V^{(1)}_{x + i\delta},
\]
and $x = x_{\min}$ denotes the smallest token in the set. 
%%%%

Similarly, the second attention score $s(z, x, t, y, n, \delta)$, which now also depends on $n, \delta$ is given by
\begin{align}
  s(z, x, t, y, n, \delta)
  \;=\;
  \mathbf{q}(z, x, n, \delta)^\top \mathbf{k}(t, y, n, \delta).\label{eq:attncontextual2}
\end{align}

%%%%
We evaluate
$n \in \{2, 3, 4\}$ and $\delta \in \{1, 5, 20\}$ for the same query token
$z = 250$ and the key-side configuration used in the main text. As we mentioned before, the second attention score depends almost entirely on $y$, namely the contextual information gathered by attention one on the key side. Since in our experiments we also fix the query token $z$, we abbreviate $s(z,x,t,y,n,\delta)$ by $s(x,t, n,\delta)$ or $s(x_{min},t, n, \delta)$.

\begin{figure}[h]
  \centering
  \includegraphics[width=\linewidth]{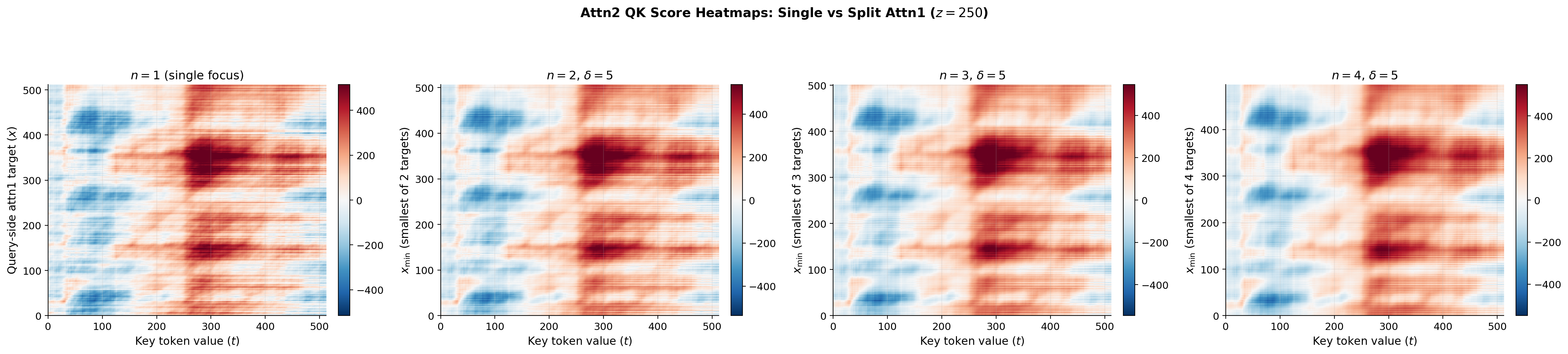}
  \caption{Second attention QK score heatmaps for $n \in \{1, 2, 3, 4\}$ equally
    attended tokens ($\delta = 5$, $z = 250$). The band structure is
    preserved across all cases.}
  \label{fig:heatmap-split-comparison}
\end{figure}

\paragraph{Heatmaps.}
Figure~\ref{fig:heatmap-split-comparison} compares the second attention's QK score heatmap
(query-side target on the $y$-axis, key token on the $x$-axis) for $n = 1$
(single focus) through $n = 4$ (four-way split, $\delta = 5$). The band
structure---elevated scores in a window above the diagonal---is preserved in
all cases. As the number of attended tokens increases, the heatmap becomes
slightly smoother but the overall geometry is unchanged.

\paragraph{Score Slices.}

\begin{figure}[h]
  \centering
  \includegraphics[width=\linewidth]{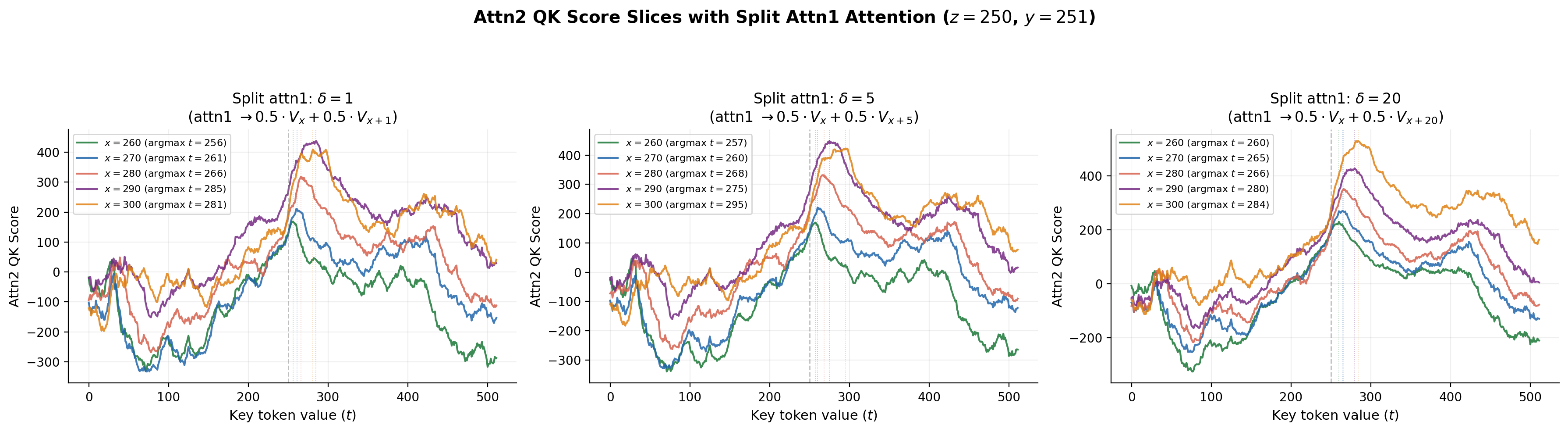}
  \caption{Second attention QK score slices with $n = 2$ equally attended tokens for
    $\delta \in \{1, 5, 20\}$.}
  \label{fig:split-attn1-slices}
\end{figure}

\begin{figure}[h]
  \centering
  \includegraphics[width=\linewidth]{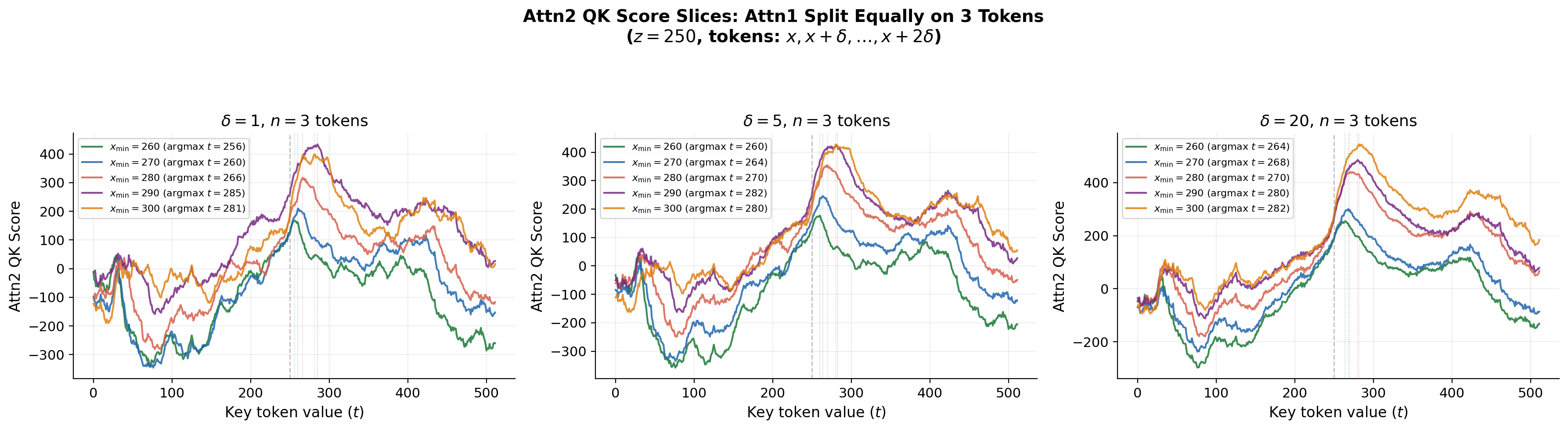}
  \caption{Second attention QK score slices with $n = 3$ equally attended tokens for
    $\delta \in \{1, 5, 20\}$.}
  \label{fig:slices-3tok}
\end{figure}

\begin{figure}[h]
  \centering
  \includegraphics[width=\linewidth]{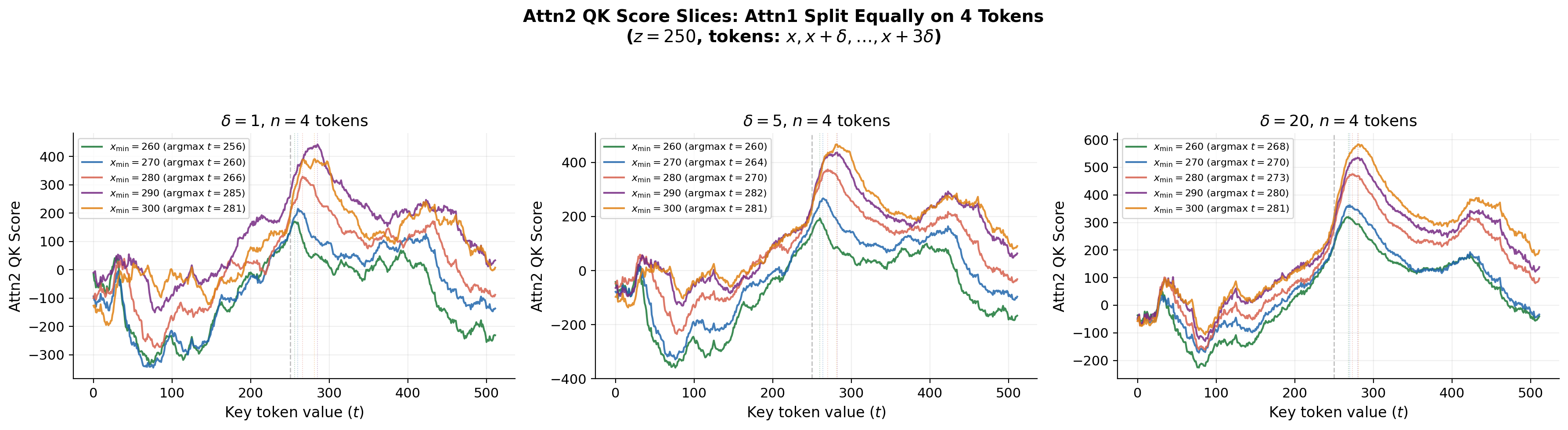}
  \caption{Second attention QK score slices with $n = 4$ equally attended tokens for
    $\delta \in \{1, 5, 20\}$.}
  \label{fig:slices-4tok}
\end{figure}

Figure~\ref{fig:split-attn1-slices} shows the second attention's QK score slices for
the two-token split ($n = 2$) at $\delta \in \{1, 5, 20\}$, with the same
query token ($z = 250$) and query-side targets
($x_{\min} \in \{260, 270, 280, 290, 300\}$) as in the main text.
Figures~\ref{fig:slices-3tok}--\ref{fig:slices-4tok} extend this to
$n = 3$ and $n = 4$. In every configuration, the band structure and
monotonicity are preserved. As $\delta$ increases, the argmax shifts slightly
toward larger key values, consistent with the larger tokens in the split
exerting a secondary upward pull on $t^*$. The effect remains modest: even at $n = 4$,
$\delta = 20$ (spanning a range of 60 tokens), the windowed monotonicity
structure is clearly preserved.

\paragraph{Argmax Bias Toward $x_{\min}$}
\label{app:argmax-bias}

When the first attention splits its weight across $n$ tokens $\{x_{\min}, x_{\min}+\delta,
\ldots, x_{\min}+(n-1)\delta\}$, the effective query-side signal is a blend of
their value vectors. A natural question is whether the argmax of the second attention's QK
score, $t^* = \arg\max_t s(x_{min}, t, n, \delta)$, tracks the minimum token $x_{\min}$, the
mean $x_{\mathrm{mean}}$, or the maximum $x_{\max}$ of the attended set.

\begin{figure}[h]
  \centering
  \includegraphics[width=\linewidth]{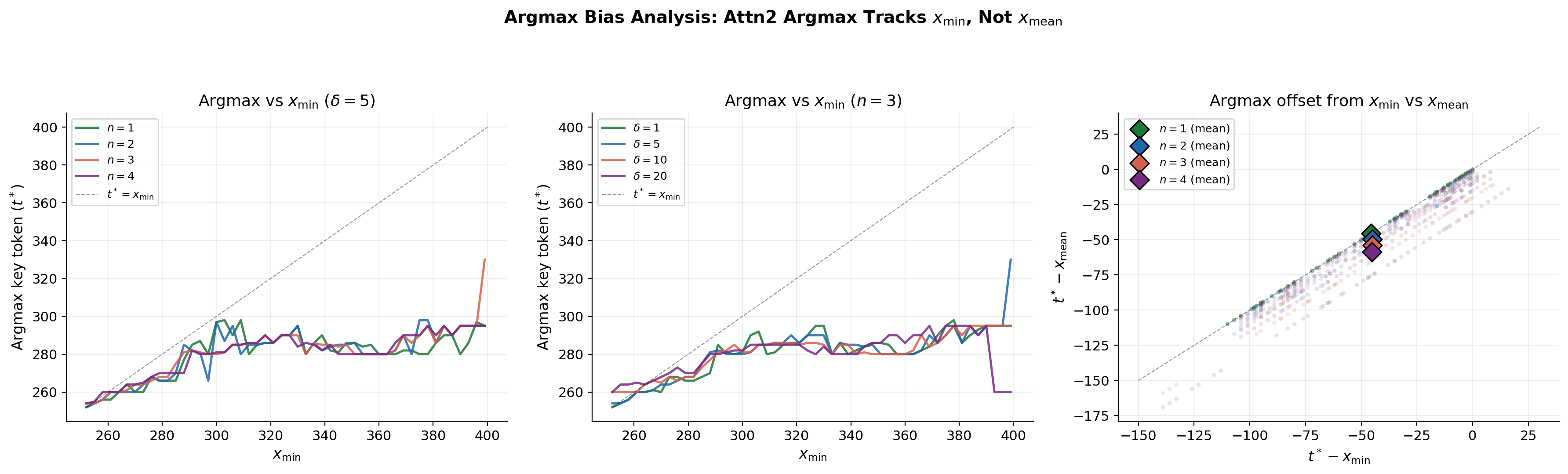}
  \caption{Argmax bias analysis ($z = 250$). \textbf{Left:} Argmax $t^*$ of the second attention vs
    $x_{\min}$ for $n \in \{1,2,3,4\}$ at $\delta = 5$. \textbf{Middle:}
    Same for $n = 3$ at varying $\delta$. \textbf{Right:} Scatter of
    $(t^* - x_{\min})$ vs $(t^* - x_{\mathrm{mean}})$ across all
    configurations; diamonds mark per-$n$ means. 
    This shows that the argmax mostly varies with the minimum of the first attention concentration set, $x_{min}$, and has almost no dependence on $n,\delta$, or $x_{mean}$. 
    In particular, the argmax consistently
    tracks $x_{\min}$, not $x_{\mathrm{mean}}$.}
  \label{fig:argmax-bias}
\end{figure}

Figure~\ref{fig:argmax-bias} provides three complementary views. The left panel
fixes $\delta = 5$ and varies $n \in \{1,2,3,4\}$: all curves track close to
$x_{\min}$ and remain well below the $t^* = x_{\min}$ diagonal (reflecting the
argmax saturation from the main text), with only minor differences between
$n$~values. The middle panel fixes $n = 3$ and varies $\delta \in \{1,5,10,20\}$:
the argmax is largely insensitive to $\delta$, confirming that the additional
higher-valued tokens in the split exert minimal upward pull on $t^*$. 
This shows that the argmax of the second attention score varies mostly as a function of $x_{min}$ and does not strongly depend on $n$ or $\delta$. 

The right
panel aggregates all $(n, \delta, x_{\min})$ configurations in a scatter of
$(t^* - x_{\min})$ versus $(t^* - x_{\mathrm{mean}})$. The mean offsets
(diamonds) show that $|t^* - x_{\min}|$ is consistently much smaller than
$|t^* - x_{\mathrm{mean}}|$ for $n > 1$, indicating that the argmax is biased
toward the smallest token in the attended set rather than its average.

This $x_{\min}$ bias is functionally advantageous for sorting. Since the first attention
identifies the next-smallest candidates above the current position's
value~$z$, the key token that the second attention should attend to is determined primarily by
the \emph{smallest} candidate, not by an average over all candidates. By
anchoring $t^*$ near $x_{\min}$, the second attention's QK mechanism effectively ignores
the distractors that the first attention also partially attends to, preserving the
conservative ``look just above $z$'' strategy described in the main text.

\section{Residual Stream Cleanup: Redundant Direct Paths}
\label{sec:residual-cleanup}

The residual stream after both layers contains five additive components:
$\mathbf{e}$ (token embedding), $\mathbf{a}_1$ (first attention output),
$\mathbf{m}_1$ (MLP1 output), $\mathbf{a}_2$ (second attention output), and
$\mathbf{m}_2$ (MLP2 output). We ask whether each component's \emph{direct}
additive contribution to the residual is necessary for the correct prediction,
or whether its information is already subsumed by downstream components that
consumed it as input.

\paragraph{Claim~1: $\mathbf{a}_1$ is redundant after feeding MLP1.}
We remove $\mathbf{a}_1$ from the residual stream immediately after it has
served as input to MLP1 (which sees
$\mathrm{LN}_2(\mathbf{e} + \mathbf{a}_1)$). The modified residual entering
block~2 is $\mathbf{e} + \mathbf{m}_1$ instead of
$\mathbf{e} + \mathbf{a}_1 + \mathbf{m}_1$; all downstream computation
(second attention, MLP2, final readout) uses this reduced residual.

\paragraph{Claim~2: $\mathbf{m}_1$ is redundant at the final readout.}
We keep $\mathbf{m}_1$ in the residual for the computation of the second attention and MLP2,
but subtract it from the final residual before applying $\mathrm{LN}_f$ and the
language model head. The readout thus sees
$\mathbf{e} + \mathbf{a}_1 + \mathbf{a}_2 + \mathbf{m}_2$ instead of the full
five-term sum.

\paragraph{Protocol.}
For each of the 40~checkpoints in our grid
(sequence lengths $\in \{16,32\}$, vocabulary sizes $\in \{128,256,512,1024\}$, 5~seeds), we first classify
the model as a two-stage model if per-token accuracy drops below 90\% when the second attention is
ablated (17 out of 40). For each two-stage model we evaluate both claims on
500~randomly sampled sequences, checking whether the argmax prediction at every
sorted-output position agrees with the unmodified forward pass.

\paragraph{Results.}
Claim~1 holds universally: across all 17~two-stage models, the argmax agreement is
$\geq 98.5\%$ (mean $99.9\%$). The argmax agreement means the argmax of the logits after removing the redundant circuit remains the same. The direct path of $\mathbf{a}_1$ through the
residual carries no decision-relevant information beyond what is already encoded
in $\mathbf{m}_1$.

Claim~2 holds for 12 of 17 two-stage models (agreement $\geq 91\%$, mean $97.6\%$),
but fails for 5 (agreement $2$--$47\%$). The failures include the two
vocabulary size $1024$ outliers previously identified in
Appendix~\ref{app:attn2-mlp1-dependence} (sequence length $16$, seed~3 and
sequence length $32$, seed~5), where the second attention exploits residual stream components
beyond $\mathbf{m}_1$, as well as three additional sequence length $16$ checkpoints
(vocabulary size $1024$, seeds~1 and~4; vocabulary size $512$, seed~5).
In these models, $\mathbf{m}_1$ carries information that is not fully captured
by $\mathbf{a}_2 + \mathbf{m}_2$ and is needed at readout.

\begin{figure}[h]
  \centering
  \includegraphics[width=\linewidth]{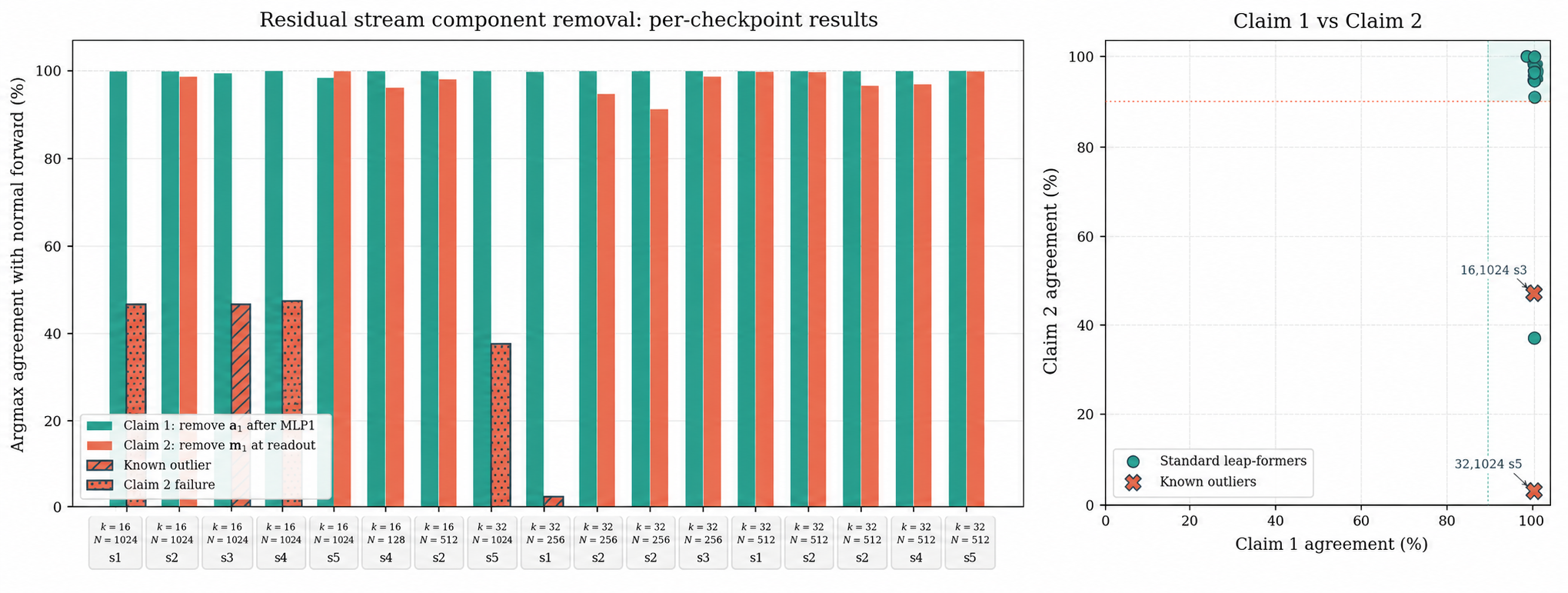}
  \caption{Residual stream component removal across 17 two-stage checkpoints.
    \textbf{Left:} Per-checkpoint argmax agreement rates for Claim~1 (remove
    $\mathbf{a}_1$, teal) and Claim~2 (remove $\mathbf{m}_1$ at readout,
    coral). The argmax agreement rate is the fraction of times removing the redundant circuit does not change the argmax of the output logits at a given token position. Hatched bars mark known outliers. \textbf{Right:} Scatter of
    Claim~1 vs Claim~2 agreement. The shaded region (${\geq}90\%$ on both axes)
    contains the 12 standard two-stage models where both direct paths are
    redundant.}
  \label{fig:residual-cleanup}
\end{figure}

Figure~\ref{fig:residual-cleanup} summarizes both claims across all
two-stage models. For the 12 standard two-stage models (excluding the 5 failures), the
model's sorting circuit can be decomposed into a clean pipeline: $\mathbf{a}_1$
feeds MLP1 and is then discarded; $\mathbf{m}_1$ feeds the second attention and MLP2 and is
then discarded; only $\mathbf{e} + \mathbf{a}_2 + \mathbf{m}_2$ is needed at
readout.

\section{Accuracy Without the Second Attention: Dependence on Current Value and Gap}
\label{sec:no-attn2-by-gap}

\begin{figure}[h]
  \centering
  \includegraphics[width=\linewidth]{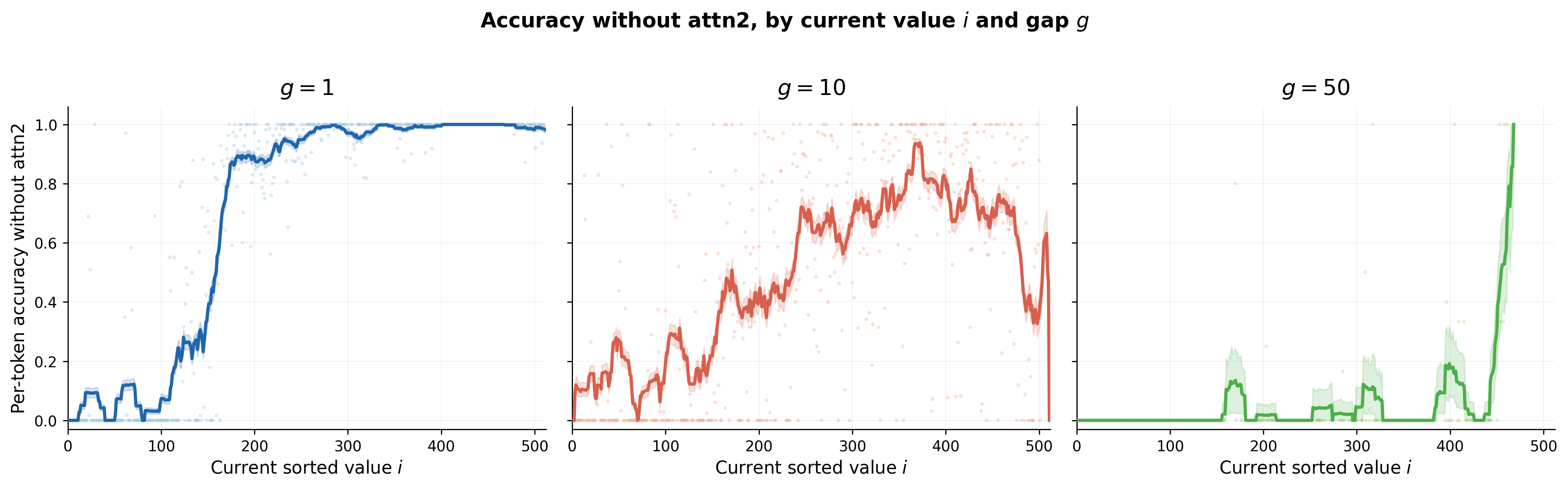}
  \caption{Per-token accuracy without the second attention as a function of current sorted
    value~$i$, for gaps $g \in \{1, 10, 50\}$ (left to right).  Grey dots
    show per-$i$ accuracy; the solid curve is a rolling average (window 20);
    shaded bands show 95\% bootstrap confidence intervals.  For small gaps
    the first-layer pathway alone achieves high accuracy once~$i$ is
    sufficiently large; for large gaps the second attention is indispensable.}
  \label{fig:no-attn2-by-gap}
\end{figure}

We ablate the second attention (skip it, retain MLP2) and measure
per-token accuracy as a function of the current sorted value~$i$ and the
gap~$g = v_{\mathrm{next}} - i$.  Figure~\ref{fig:no-attn2-by-gap} shows
three representative gaps: $g = 1$, $g = 10$, and $g = 50$.

For the smallest gap ($g = 1$), the model without the second attention is nearly at chance
when~$i$ is small, but accuracy rises steeply around $i \approx 150$ and
saturates close to~$1$ for $i \gtrsim 250$.  At $g = 10$ a similar ramp
appears but saturates at a lower level (${\approx}0.7$), and at $g = 50$
accuracy remains near zero for all values of~$i$.

This demonstrates that the first-layer pathway alone can resolve the correct
next token only when the target lies close in value to the current
token---a regime where the token embedding alone provides
enough information without the fine-grained windowed search that the second attention
performs.  As the gap grows, the second attention becomes indispensable.

\section{Corroboration of Gap-Dependent Specialization via Hijacking}
\label{app:hijacking}

This appendix elaborates on the hijacking intervention introduced in the main
text (\S\ref{sec:specialization}), which we use to causally corroborate the
division of labor across gap sizes. Specifically, for a given input sequence
and output position, we independently redirect the direct circuit of the first
attention and the attention focus of the second attention toward a different
token in the sequence, and measure how often each intervention alone, or both
simultaneously, succeeds in steering the model's prediction to the redirected
token.

Figure~\ref{fig:hijack-fl-avg} reports the results, averaged over 23~two-stage
seeds on the configuration with sequence length $32$ and vocabulary size $512$,
with shaded bands showing $\pm 1$ standard deviation across seeds. Each panel
corresponds to a different minimum gap~$g$ between the current sorted value and
the next correct token, and the $x$-axis varies the offset of the hijack target
(i.e., the redirected token is the one at position $i + $\emph{offset} in the
vocabulary).

When the gap is small ($g = 1$), both the direct-circuit hijack of the first
attention (red) and the second attention hijack (blue) individually succeed at rates which vary across seeds. The purple curve shows the fraction of instances in which
\emph{both} interventions independently steer the prediction to the hijacked
token on the same input---at gap~$1$, this co-success rate is approximately
$20\%$ on average. This indicates that the final output logit can be shaped by both the direct
circuit of the first attention and the output circuit of the second attention
simultaneously, akin to a voting mechanism in which the two pathways jointly
determine the predicted token.

As the gap increases, the direct circuit of the first attention rapidly loses
influence: at $g = 10$ the red curve has already declined substantially, and by
$g = 20$ the direct-circuit hijack success rate is near zero. In contrast, the
second attention hijack (blue) remains close to $100\%$ across all gap sizes,
confirming that for larger gaps the second attention is the dominant, and
eventually sole, circuit responsible for specifying the next token. The
transition from shared influence at small gaps to second attention dominance at
large gaps is consistent with the ablation and attention-mass results described
in the main text, providing causal, intervention-based confirmation of
gap-dependent specialization.

\end{document}